\documentclass{article}[camera ready] 
\usepackage{iclr2024_conference,times}
\usepackage{comment}
\usepackage[colorlinks=true, allcolors=blue]{hyperref}
\usepackage{url}
\usepackage{amsmath,amsfonts,amssymb}
\usepackage{algorithm}
\usepackage{algpseudocode}
\usepackage{graphicx}
\usepackage{bm}
\usepackage[english]{babel}
\usepackage{amsthm}
\usepackage{bbm}
\usepackage[normalem]{ulem}
\usepackage{enumitem}
\usepackage{xcolor}
\newcommand{\prk}[1]{{[\color{black}PRK: #1}]}
\newcommand{\pch}[1]{\textcolor{black}{#1}}

\newcommand{\hungyh}[1]{\textcolor{black}{#1}}

\DeclareMathOperator*{\argmax}{argmax} 

\DeclareMathOperator{\btheta}{\bm{\theta}}
\DeclareMathOperator{\bthetaR}{\btheta^{\textbf{V}}_t}

\DeclareMathOperator{\bA}{\mathbf{A}}

\DeclareMathOperator{\bI}{\mathbf{I}}
\DeclareMathOperator{\cS}{\mathcal{S}}
\DeclareMathOperator{\cA}{\mathcal{A}}
\DeclareMathOperator{\cF}{\mathcal{F}}
\newtheorem{assumption}{Assumption}
\newtheorem{theorem}{Theorem}
\newtheorem*{theorem*}{Theorem}
\newtheorem{lemma}{Lemma}
\newtheorem*{lemma*}{Lemma}

\newtheorem{remark}{Remark}
\newtheorem{corollary}{Corollary}
\newcommand{\innerprod}[2]{\langle#1,#2 \rangle}
\newcommand{\MLE}[1]{\btheta^{\normalfont\text{MLE}}_{#1}}
\newcommand{\CLT}[1]{\frac{37d^2}{p^2_{\text{min}}}\cdot \log\left( \frac{d\lambda+#1L^2}{d}\right)\cdot \log{\frac{1}{\delta}}}

\title{Value-Biased Maximum Likelihood Estimation for Model-based Reinforcement Learning in Discounted Linear MDPs}

\author{Yu-Heng Hung \& Ping-Chun Hsieh \\
Department of Computer Science\\
National Yang Ming Chiao Tung University\\
Hsinchu, Taiwan \\
\texttt{\{hungyh.cs08,pinghsieh\}@nycu.edu.tw} \\
\And
Akshay Mete \& P. R. Kumar \\
Texas A\&M University \\
College Station, Texas, USA \\
\texttt{\{akshaymete,prk\}@tamu.edu} 
}

\iclrfinalcopy 
\begin{document}

\maketitle

\begin{abstract}
We consider the infinite-horizon linear Markov Decision Processes (MDPs), where the transition probabilities of the dynamic model can be linearly parameterized with the help of a predefined low-dimensional feature mapping. While the existing regression-based approaches have been theoretically shown to achieve nearly-optimal regret, they are computationally rather inefficient due to the need for a large number of optimization runs in each time step, especially when the state and action spaces are large.
To address this issue, we propose to solve linear MDPs through the lens of Value-Biased Maximum Likelihood Estimation (VBMLE), which is a classic model-based exploration principle in the adaptive control literature for resolving the well-known closed-loop identification problem of Maximum Likelihood Estimation. We formally show that (i) VBMLE enjoys $\widetilde{O}(d\sqrt{T})$ regret, where $T$ is the time horizon and $d$ is the dimension of the model parameter, and (ii) VBMLE is computationally more efficient as it only requires solving one optimization problem in each time step. In our regret analysis, we offer a generic convergence result of MLE in linear MDPs through a novel supermartingale construct and uncover an interesting connection between linear MDPs and online learning, which could be of independent interest. Finally, the simulation results show that VBMLE significantly outperforms the benchmark method in terms of both empirical regret and computation time.





\end{abstract}

\section{Introduction}
Model-based reinforcement learning (MBRL) is one fundamental paradigm that learns an optimal policy by alternating between two subroutines: estimation of the transition dynamics and planning according to the learned dynamics model.
MBRL has been extensively studied in the tabular setting from various perspectives, including \citep{auer2008near, azar2017minimax}, which have been shown to achieve either optimal regret bounds or sample complexity.
Despite the above success, the conventional tabular MBRL methods are known to be computationally intractable in RL problems with large state or action spaces due to the need for direct estimation and access to the per-state transition probability.
To enable MBRL for large state and action spaces, one important recent attempt is to study Markov decision processes with linear feature mappings \citep{zhou2021provably}, which is termed linear MDPs subsequently in this paper. Specifically, linear MDPs assume that the probability of each transition can be represented by $\langle \phi(s'|s,a), \btheta^{*} \rangle$, where $\phi(\cdot|\cdot,\cdot)$ is a known feature function for each possible transition, and $\btheta^{*}$ parametrizes the transition probabilities to be learned. This framework can readily encompass various related formulations, such as tabular MDPs, feature-based linear transition models \citep{yang2019sample}, the linear combination of base models \citep{modi2020sample}, and linear value function frameworks \citep{zanette2020learning}.

Based on the existing literature of linear MDPs, the existing approaches could be divided into two primary categories depending on the length of the learning horizon: \textit{episodic MDPs} and \textit{infinite-horizon discounted MDPs}. In episodic MDPs, one important feature is that the environment state could be conveniently reset to some initial state when a new episode starts. Several recent works have explored episodic MDPs through the use of value-targeted regression techniques, e.g., \citep{ayoub2020model,zhou2021nearly}. 
A more detailed survey of the related works for episodic linear MDPs is deferred to Section \ref{sec:related}.
By contrast, in infinite-horizon linear MDPs, due to the absence of the periodic restart, addressing exploration and conducting regret analysis could be even more challenging than that in episodic MDPs. 
Some recent attempts tackles infinite-horizon linear MDPs by designing regression-based approaches and establishing theoretical guarantees \citep{zhou2021nearly, zhou2021provably, chen2022sample}. 
However, their algorithms could suffer from high computational complexity and be intractable in practice for the following reasons: (i) In the existing regression-based approaches, in each time step, one would need to solve a constrained optimization problem of the action-value function for \textit{each} state-action pair. (ii) Moreover, in order to represent the value function as a linear combination of the learned parameter vector $\btheta$, it is necessary for those regression-based approaches to construct a vector $\phi_{V}(s,a) := \sum_{s'\in\mathcal{S}} \phi(s'|s,a)V(s')$. As a result, the action-value function can be expressed as follows: $Q(s,a) = \innerprod{\phi_{V}(s,a)}{\btheta}$. However, constructing $\phi_V$ could be computationally intractable when dealing with a large state space. These limitations render the regression-based approaches mentioned above rather challenging to implement and deploy in practice.
Therefore, one important research question remains to be answered: \textit{How to design an efficient model-based RL algorithm for infinite-horizon discounted linear MDPs with provable regret guarantees?}

In this paper, we answer the above question affirmatively. Specifically, to address the above limitations, we design a tractable approach based on the classic principle of Value-Biased Maximum Likelihood Estimation (VBMLE) \citep{kumar1982optimal}, which has shown promising results in recent developments in bandits \citep{hung2021reward,hung2023reward} and tabular RL \citep{mete2021reward}, and leverage the value biasing technique to enforce exploration. 
The major advantage of VBMLE is that with the help of value biasing, it requires solving only one optimization problem for learning the dynamics model parameter at each time step and thereby enjoys a significantly lower computational complexity than the regression-based approaches. 
Moreover, we formally establish $\tilde{\mathcal{O}}({d\sqrt{T}})$ regret bound based on the following novel insights: (i) We establish a convergence result on the Maximum Likelihood Estimator for linear MDPs by using a novel supermartingale approach. (ii) Through this construct, we also find useful connections between (i) the linear MDPs and online portfolio selection problem as well as (ii) VBMLE and the Follow-the-Leader algorithm in online learning. We highlight the main contributions as follows:

\begin{itemize}[leftmargin=*]
    \item We adapt the classic VBMLE principle to the task of learning the dynamic model for linear MDPs. Our proposed algorithm addresses model-based RL for linear MDPs from a distributional perspective, which learns the parameterized transition directly by maximum likelihood estimation without resorting to regression, and guides the exploration via value biasing instead of using concentration inequalities.
    \item We establish the theoretical regret bound of VBMLE by providing a novel theorem connected to the confidence ellipsoid of MLE. Furthermore, we uncover an interesting connection between online learning and our regret analysis.
    \item We conduct an empirical analysis to assess both the computational complexity and empirical regret performance. The simulation results demonstrate that VBMLE exhibits a clear advantage in terms of both effectiveness in regret and computational efficiency.
    \end{itemize}

\section{Related Works}
\label{sec:related}
\textbf{VBMLE for Multi-Armed Bandits and RL}.
Regarding VBMLE, various prior works have applied this method to different bandit settings and tabular MDP.
Firstly, \citep{liu2020exploration} focuses on solving non-contextual bandits with exponential family reward distributions.
Next, \citep{hung2021reward} introduces two variations of VBMLE: LinRBMLE and GLM-RBMLE. These methods are designed for solving linear contextual bandits and result in an index policy.
Furthermore, \citep{hung2023reward} leverages the representation power of neural networks and proposes NeuralRBMLE. This approach is specifically designed for solving neural bandits, making no assumptions about the unknown reward distribution. As for the MDP setting, \citep{mete2021reward} has adapted VBMLE to solve tabular MDPs, where the states and actions belong to a known finite set, while \citep{mete2022augmented} analyzed the finite performance of a constrained version of VBMLE. 
By contrast, this paper takes the very first step towards understanding the theoretical regret performance of VBMLE in RL beyond the tabular settings.


\textbf{Episodic Linear MDPs}. Denoting $H$ is the total episode considered, \citep{cai2020provably} has applied Proximal Policy Optimization (PPO). \citep{ayoub2020model} utilizes value target regression as an optimism principle.
\citep{zhou2021nearly} provides a new tail inequality and adapts weighted ridge regression on UCRL-VTR provided by \cite{ayoub2020model}. All of their algorithm achieve a regret bound of $\mathcal{O}(d\sqrt{H})$. On the other hand, \citep{jin2020provably} provides a model-free approach that considers the upper confidence bound on the action-value function. \citep{wang2019optimism,wang2020reinforcement} presents a value-based approach for MDPs with generalized linear function approximation but requires an optimistic closure assumption, and all of them have $\mathcal{O}(\sqrt{d^3H})$ regret bound.

\textbf{Infinite-Horizon Discounted Linear MDPs}. In the context of infinite-horizon discounted linear MDPs, \citep{zhou2021provably} introduced the UCLK algorithm. This algorithm takes into consideration the confidence set on the least-square estimator of $\btheta^*$ and demonstrates a regret upper bound of $\mathcal{O}(d\sqrt{T}/(1-\gamma)^2)$. \citep{zhou2021nearly} introduced an improved version of the UCLK algorithm, called $\text{UCLK}^+$. $\text{UCLK}^+$ incorporates weighted ridge regression into the original UCLK algorithm and achieves a regret bound that matches the established lower bound of $\mathcal{O}(d\sqrt{T}/(1-\gamma)^{1.5})$. On the other hand, \citep{chen2022sample} also provide a variant of the UCLK algorithm, called UPAC-UCLK, which has $\mathcal{O}(d\sqrt{T}/(1-\gamma)^2) + \mathcal{O}(\sqrt{T}/(1-\gamma)^3)$ regret bound but with uniform-PAC sample complexity guarantee.

\section{Problem Formulation}
\label{section:problem setting}
\textbf{Markov Decision Processes (MDP) and Linear Feature Mapping.} An MDP is denoted by $\mathcal{M}:=\langle \mathcal{S}, \mathcal{A}, P, {R}, T, \mu_0\rangle$, where $\mathcal{S}$ and $\mathcal{A}$ 
represent the state and action spaces, respectively, $P$ is the dynamic model, $R: \mathcal{S} \times \mathcal{A} \rightarrow [0,1]$ is the reward function, $T$ is the time horizon, and $\mu_0$ is the initial state distribution with $\mu_0(s)>0$\footnote{As there is a policy that achieves optimal value for \textit{all} initial states $s$, or equivalently, for \textit{all} initial distributions $\mu_0$, without loss of generality, it is common to take a strictly positive initial distribution.}. 
A linear MDP is defined by the following: 
\begin{itemize}[leftmargin=*]
    \item There exist an unknown parameter $\btheta^{*} \in \mathbb{R}^{d}$, and a known feature mapping $\phi(\cdot|\cdot,\cdot): \cS\times \cA\times \cS \rightarrow \mathbb{R}^d$, such that $P(s'|s,a) = \innerprod{\phi(s'|s,a)}{\btheta^{*}}, \forall s',s \in \mathcal{S}, a \in \mathcal{A}$. 
    \item $\lVert\btheta^*\rVert_2 \leq \sqrt{d}$ and $\lVert\phi(s'|s,a)\rVert_2 \leq L, \forall s',s \in \mathcal{S}, a \in \mathcal{A} $.
\end{itemize}

Moreover, let $\mathbb{P}$ denote the set of parameters that correspond to the product of the simplices for each (state, action) pair: 
\begin{align}
    \mathbb{P} := \left\{ \btheta : 0 \leq \langle \phi(\cdot | s, a), \btheta \rangle \leq 1, \sum_{s' \in \mathcal{S}} \langle \phi(s' | s, a), \btheta \rangle = 1 , \forall s \in \mathcal{S}, a\in \mathcal{A}\right\},
\end{align}
where $\btheta$ denotes the parameter of the transition dynamics model and $\phi(\cdot \rvert s, a)$ is the known feature mapping function.

A policy $\pi: \mathcal{S}\rightarrow \Delta(\mathcal{A})$, where $\Delta(\mathcal{A})$ is the set of all probability distributions on $\mathcal{A}$, designed to maximize the sum of discounted reward, which is denoted by the value function:
\begin{align}
    V^{\pi}(s;\btheta) := \mathbb{E}_{\substack{a_{i} \sim \pi(\cdot|s_{i}) \\ s_{i+1}\sim \langle \phi(\cdot|s_{i},a_{i}).\btheta\rangle}} \left[ \sum_{i=0}^{\infty}\gamma^{i}r(s_{i},a_{i})\Big|s_0 = s\right].
\end{align}
Similarly, the action value function $Q^{\pi}(s, a;\btheta)$ is defined as
\begin{align}
    Q^{\pi}(s,a;\btheta) := \mathbb{E}_{\substack{a_{i} \sim \pi(\cdot|s_{i}) \\ s_{i+1}\sim \langle \phi(\cdot|s_{i},a_{i}).\btheta\rangle}} \left[ \sum_{i=0}^{\infty}\gamma^{i}r(s_{i},a_{i})\Big|s_0 = s, a_0 = a \right].
\end{align}
Moreover, we let $J(\pi;\btheta) := \mathop{\mathbb{E}}_{s\sim\mu_0}[V^*(s;\btheta^{*})]$ denote the mean reward achievable for the MDP with parameter $\btheta$ under policy $\pi$ over the initial probability distribution $\mu_0$. 

\textbf{Optimal Value and Regret.}
We then define the optimal value function to be the maximum value obtained by a policy: $V^*(s;\btheta) = \max_{\pi} V^{\pi}(s;\btheta)$. In the discounted linear MDP setting \citep{zhou2021provably}, the cumulative regret $\mathcal{R}(T)$ for the MDP with parameter $\btheta^*$ is defined to be the total difference of value function between the optimal policy and the learned policy $\pi_t$, where
\begin{align}
    \mathcal{R}(T):= \sum_{t=1}^{T} \left[V^*(s_t;\btheta^{*}) - V^{\pi_t}(s_t,\btheta^*)\right], \quad  s_1 \sim \mu_0. \label{eq: def of regret}
\end{align}

Based on the fundamental result that there exists a policy that achieves optimal value for \textit{all} states, we use $\pi^{*}(\btheta)$ to denote an optimal policy with respect to a given model parameter $\btheta$ as
\begin{align}
    \pi^{*}(\btheta) := \argmax_{\pi} J(\pi; \btheta).
\end{align}

\section{VBMLE for Linear MDPs}


\textbf{Introduction to the VBMLE Principle.} We now introduce the idea behind the classic value biasing principle.
Consider first the certainty equivalence principle \citep{kumar2015stochastic} employing the straightforward \hungyh{ Maximum Likelihood Estimate (MLE) as
\begin{align}
    \widehat{\btheta}_t := \argmax_{\btheta\in\mathbb{P}} \left\{\prod_{i=1}^{t-1} p(s_{i+1}|,s_i,a_i;\btheta) \right\}, \label{MLE}
\end{align}
}That is, at each time step $t$, the learner employs the policy $\pi_{t}^{\text{MLE}}$ that is optimal for the current estimate $\widehat{\btheta}_t$. Under appropriate technical conditions, it has been shown in \citep{borkar1979adaptive} that $\widehat{\btheta}_t$ converges almost surely to a random 
$\widehat{\btheta}_\infty$, for which
\begin{align}
    p(s'| s,\pi_{\infty}^{\text{MLE}}(s);\widehat{\btheta}_\infty) = p(s'| s,\pi_{\infty}^{\text{MLE}}(s);\btheta^*), 
    \forall s, s' \in \mathcal{S}.
    \label{eq: close loop}
\end{align}
This convergence property is called the \textit{closed-loop identification} property; it means that asymptotically the transition probabilities resulting from the application of the
policy $\pi_{\infty}^{\text{MLE}}$ are correctly estimated. An important consequence is that
$J(\pi^{*}(\widehat{\btheta}_\infty);\widehat{\btheta}_\infty) = J(\pi^{*}(\widehat{\btheta}_\infty);\btheta^*)$.
Since $\pi_{\infty}^{\text{MLE}}$ is not necessarily optimal for $\btheta^*$, this implies that
\begin{align}
    J(\pi^{*}(v);\widehat{\btheta}_\infty) \leq J(\pi^{*}(\btheta^*);\btheta^*). \label{eq: value bias}
\end{align}
The idea of the Value-Biased method is to try to undo the bias in (\ref{eq: value bias}) by adding a bias term that favors parameters with larger optimal total return. This leads to the principle of Value-Biased Maximum Likelihood Estimate (VBMLE) originally proposed in the adaptive control literature by \citep{kumar1982optimal} as follows:
\begin{align}
    \btheta^{\normalfont\textbf{VBMLE}}_t := \argmax_{\btheta \in \mathbb{P}} \left\{ \sum\limits_{i=1}^{t-1} \log p(s_{i+1}|s_i,a_i;\btheta) + \alpha(t)\cdot J(\pi^{*}(\btheta);\btheta)  \right\},
\end{align}
where $\alpha(t)$ is a positive increasing sequence that weights the bias in favor of parameters with larger total return. VBMLE employs this biasing method to handle the exploration-exploitation trade-off.

\textbf{VBMLE for Discounted Linear MDPs.} 
In this paper, we adapt the VBMLE principle to the RL problem in the linear MDP setting. Specifically, at each time step, the learner would (i) choose the parameter estimate that maximizes the regularized log-likelihood plus the value-bias as
\begin{align}
    \bthetaR := \argmax_{\btheta \in \mathbb{P}} \left\{ \sum\limits_{i=1}^{t-1} \log\langle \phi(s_{i+1}|s_i,a_i),\btheta \rangle  + \frac{\lambda}{2}\lVert \btheta\rVert_2^2  + \alpha(t)\cdot V^*(s_t;\btheta)  \right\},\label{eq:VBMLE theta}
\end{align}
where $\lambda$ is a positive constant for regularization, and then (ii) employ an optimal policy with respect to $\bthetaR$. Notice that the term $V^*(s_t;\btheta)$ can be computed by using the standard Value Iteration presented as Algorithm \ref{alg:vi} in Appendix. 
If there are multiple maximizers for (\ref{eq:VBMLE theta}), then one could break the tie arbitrarily.
For clarity, we also summarize the procedure of VBMLE in Algorithm \ref{alg:VBMLE}.

\begin{algorithm}[!t]
\caption{VBMLE for Reinforcement Learning in Linear MDPs}
    \begin{algorithmic}[1]
        \State {\bfseries Input:} $\alpha(t)$
        \For{$t=1,2,\cdots$}
            \State $\bthetaR := \argmax_{\btheta \in \mathbb{P}} \left\{ \sum\limits_{i=1}^{t-1} \log\langle \phi(s_{i+1}|s_i,a_i),\btheta \rangle  + \frac{\lambda}{2}\lVert \btheta\rVert_2^2 + \alpha(t)\cdot V^*(s_t;\btheta)  \right\}$.
            \State $a_t = \argmax\limits_{a \in \mathcal{A}}Q^*(s_t,a;\btheta^{\textbf{R}}_t)$ 
        \EndFor
    \end{algorithmic}
    
\label{alg:VBMLE}
\end{algorithm}

\textbf{Features of VBMLE for Linear MDPs.} We highligh the salient features of the VBMLE method in Algorithm \ref{alg:VBMLE} as follows.
\begin{itemize}[leftmargin=*]
    \item \textbf{Computational Efficiency:} As mentioned earlier, UCLK \citep{zhou2021provably} suffers from high computational complexity as it requires computing an estimate of the model parameter for \textit{each} state-action pair in each iteration. This renders UCLK intractable when either the state space or the action space is large. By contrast, the VBMLE approach, which applies value-bias to guide the exploration under the MLE, only requires solving one single maximization problem for the dynamics model parameter $\btheta$ in each iteration, making it computationally efficient and superior. Accordingly, VBMLE could serve as a more computationally feasible algorithm for RL in linear MDPs in practice. 
    \item \textbf{VBMLE is Parameter-Free:} As shown in Algorithm \ref{alg:VBMLE}, the only parameter required by VBMLE is $\alpha(t)$, which determines the weight of the value bias. As will be shown in Section \ref{sec:regret}, one could simply choose $\alpha(t)=\sqrt{t}$ to achieve the required regret bound, and moreover this simple choice also leads to superior empirical regret performance. As a result, VBMLE is parameter-free and therefore does not require any hyperparameter tuning.
    \item \textbf{Distributional Perspective:} In contrast to the existing RL methods for linear MDPs \citep{ayoub2020model, zhou2021nearly, zhou2021provably, chen2022sample} that aim to learn the unknown parameter via regression on the value function (or termed \textit{value-targeted regression}), the proposed VBMLE takes a distributional perspective through directly learning the whole collection of transition probabilities through value-biased maximum likelihood estimation. 
    This perspective has also been adopted by the prior works on applying VBMLE to the contextual bandit problems \citep{hung2021reward,hung2023reward}. 

\end{itemize}

\textbf{Differences Between VBMLE for RL and VBMLE for Bandits.} Compared to the existing works on VBMLE for bandits \citep{hung2021reward,hung2023reward}, VBMLE for RL presents its own salient challenges:
\begin{itemize}[leftmargin=*]
\item \textbf{Non-Concave Objective Function of VBMLE for Linear MDPs:} In the context of bandits, VBMLE learns by maximizing the log-likelihood of observed \textit{rewards} with a bias term that depends on the maximum achievable reward. As the parametric form of the reward distributions is typically unknown in the bandit setting, \citep{hung2021reward,hung2023reward} rely on a surrogate likelihood function (typically belongs to an exponential family) to estimate the unknown reward distributions and incorporates a reward bias by adding the immediate maximum reward. As a result, the resulting objective function still remains a concave function such that its maximizer either enjoys a closed-form expression or could be solved efficiently by a gradient-based method. By contrast, VBMLE for RL manages to optimize the log-likelihood with the value bias, which ends up as a \textit{non-concave} function. 
To address this issue, we take the following approach: (i) For the theoretical regret analysis, we consider an oracle that returns the maximizer of the constrained optimization problem induced by VBMLE. (ii) For the practical implementation, we could incorporate the value iteration into the optimization subroutine and use a gradient-based method to numerically find an approximate maximizer of VBMLE.
\item \textbf{Non-Index-Type Policies:} Prior works on VBMLE for bandits \citep{liu2020exploration,hung2021reward,hung2023reward} could convert the original VBMLE into an \textit{index-type policy} by using arm-specific estimators. 
This conversion also facilitates the regret analysis in \citep{liu2020exploration,hung2021reward,hung2023reward}.
However, this approach is not applicable in RL for linear MDPs since the action and state spaces could typically be very large in practice. 
As a result, we are not allowed to reuse a similar analytical framework to characterize the regret of VBMLE in linear MDPs.
To address this, we leverage a supermartingale approach and use an induction argument to establish the regret bound, as will be shown in Section \ref{sec:regret}.
\end{itemize}

\section{Regret Analysis}
\label{sec:regret}
In this section, we formally present the regret analysis of the VBMLE algorithm. To begin with, we introduce the following useful notations:

\begin{align}
    \ell_t(\btheta) :=& \sum\limits_{i=1}^{t-1} \log \innerprod{\phi(s_{i+1}|s_i,a_i)}{\btheta} + \frac{\lambda}{2}\lVert \btheta \rVert_2^2 \\
    \MLE{t} :=& \argmax\limits_{\btheta \in \mathbb{P}} \ell_t(\btheta), \label{eq:theta_MLE} \\
    \bA_t :=& \sum_{i=1}^{t-1}\phi(s_{i+1}|s_i,a_i)\phi(s_{i+1}|s_i,a_i)^\top +\lambda I. \label{eq:A_t}
\end{align}
If there are multiple maximizers for (\ref{eq:theta_MLE}), then one could break the tie arbitrarily.

\begin{assumption}
    \label{assumption: p_min}
    The following information for the transition probability $P$ is known:
    \begin{itemize}
        \item The set of zero transition $P_{0} := \{ (s,a,s')| P(s'|s,a) = 0, \forall s,s' \in \mathcal{S}, a \in \mathcal{A}\}$.
        \item The lower bound non-zero transition probabilities $p_{\text{min}} := \min_{(s,a,s') \not\in P_{0}} P(s'|s,a)$.
    \end{itemize}
    We then redefine the probability simplex 
 based on the above assumption as follows:
    \begin{align}
        \mathbb{P} := \bigg\{ \btheta \bigg| & \;p_{\text{min}} \leq \innerprod{\phi(s' | s, a)}{\btheta} \leq 1, \forall (s,a,s') \not\in P_0; \nonumber \\
        &\;\innerprod{\phi(s' | s, a)}{\btheta} = 0, \forall (s,a,s') \in P_0; \nonumber \\
        &\; \sum_{s' \in \mathcal{S}} \innerprod{\phi(s' | s, a)}{\btheta} = 1 , \forall s \in \mathcal{S}, a\in \mathcal{A}\bigg\}.\label{simplex}
    \end{align}
\end{assumption}
\begin{remark}
    \normalfont This assumption suggests that the magnitude of the gradient of the log probability for the observed transition, denoted as $\lVert \nabla_{\btheta} \log{\innerprod{\phi(s_{i+1}|s_i,a_i)}{\MLE{t}}} \rVert_2$, is bounded from above. A similar assumption is made in \citep{kumar1982optimal,mete2021reward}.
\end{remark} 

\subsection{Convergence Analysis of MLE in Linear MDPs}
\newcommand{\SVN}[1]{2d\log\left( \frac{d\lambda+#1L^2}{d}\right)}
To begin with, we highlight the main technical challenges as follows: A natural idea is to leverage the Azuma–Hoeffding inequality on the log-likelihood ratio: $\ell_{t}(\MLE{t}) - \ell_{t}(\btheta^{*})$, and then find the distance between $\MLE{t}$ and the true parameter $\btheta^*$. However, it is known that the stochastic process induced by the maximum log-likelihood ratio is actually a \textit{sub-martingale} (shown in Lemma \ref{lemma: submartingale} in Appendix for completeness).
To address these issue, we propose several novel techniques: (i) We first propose to construct a novel super-martingale (cf. Lemma \ref{lemma: supermartingale}) to characterize the convergence rate of the MLE in linear MDPs, which could be of independent interest beyond RL problems. Interestingly, this supermartingale consists of a term that could be interpreted as the regret in the online portfolio selection problem and thereby offers an interesting connection between linear MDPss and online learning. (ii) Built on (i), to utilize Azuma-Hoeffding inequality, we need to carefully handle the sum of squared supermartingale differences, which do not have an explicit uniform upper bound and require a more sophisticated argument.


\begin{lemma}[Lemma C.2 in \citep{zhou2021provably}, Lemma 11 in \citep{abbasi2011improved}]
    \label{lemma: summation of vector norm}
    Given any $\{\phi(s_{t+1}|s_t,a_t) \}_{t=1}^T \in \mathbb{R}^d$ satisfying that $\lVert \phi(s_{t+1}|s_t,a_t)\rVert_2 \leq L$. For all $\lambda > 0$, we have
    \begin{align}
        \sum_{t=1}^T \lVert \phi(s_{t+1}|s_t,a_t)\rVert^2_{\bA^{-1}_t} \leq \SVN{T}. \label{lemma: summation of vector norm eq-1}
    \end{align}
\end{lemma}

\begin{lemma}
\label{Lemma: upper bound of V}
$\forall s \in \mathcal{S},a \in \mathcal{A}, \text{and } \btheta \in \mathbb{P}$, we have $Q^*(s,a,\btheta) \leq \frac{1}{1-\gamma}$.
\end{lemma}
Lemma \ref{Lemma: upper bound of V} is a direct result of that the reward function is bounded by 1, i.e., $R(\cdot,\cdot)\leq 1$.

We proceed to construct two useful helper stochastic processes as follows: For each $t\in \mathbb{N}$,
\begin{align}
      X_t &:= \ell_{t}(\MLE{t-1}) - \ell_{t}(\btheta^{*}) + \sum_{i=1}^{t-1} z_i,\\
      z_t &:= \ell_{t}(\MLE{t}) - \ell_{t}(\MLE{t-1}).
\end{align}

\begin{lemma}[Azuma–Hoeffding Inequality]
    \label{lemma: Azuma–Hoeffding inequality}
    Suppose $\{X_1,X_2,\cdots\}$ is a martingale or super-martingale. Then for all positive integers $t$ and all positive reals $\delta$, we have
    \begin{align}
    \normalfont
        \text{Pr}\left(X_t - X_0 \geq \sqrt{2M_t\log{\frac{1}{\delta}}}\right) \leq \delta, 
    \end{align}
    where $M_t := \sum_{i=1}^{t} (X_t-X_{t-1})^2$.
\end{lemma}
\hungyh{
\begin{lemma}
    \label{lemma: martingale}
  For all $\lambda \geq 0$, the stochastic process $\{L_{t}(\MLE{t-1}) \cdot \prod_{i=1}^{t-1}\text{\normalfont exp}(-z_i)\}$ is a martingale, i.e.,
    \begin{align}
        \normalfont
        \mathbb{E}_{s_{t+1}\sim \text{Pr}(\cdot|s_t,a_t;\btheta^{*})}\left[ L_{t+1}(\MLE{t})\cdot\prod_{i=1}^{t}\text{exp}(-z_i)\bigg\rvert\mathcal{F}_t \right] = L_{t}(\MLE{t-1}) \cdot \prod_{i=1}^{t-1}\text{exp}(-z_i), \label{lemma: martingale eq-0}
    \end{align}
    where $\mathcal{F}_{t}:= \{s_1,a_1,\cdots,s_{t},a_{t}\}$ denotes the causal information up to time $t$.  
\end{lemma}
\begin{corollary} 
    \label{lemma: supermartingale}
  For all $\lambda \geq 0$, the stochastic process $\{X_t\}$ is a supermartingale, i.e.,
    \begin{align}
        \normalfont
        \mathop{\mathbb{E}}_{s_{t+1} \sim \text{Pr}(\cdot|s_t,a_t;\btheta^*)} \left[\ell_{t+1}(\MLE{t}) - \ell_{t+1}(\btheta^{*}) - \sum_{i=1}^{t}z_i\bigg| \mathcal{F}_t\right] \leq \ell_{t}(\MLE{t-1}) - \ell_{t}(\btheta^{*}) - \sum_{i=1}^{t-1}z_i.  \label{lemma: supermartingale eq-0}
    \end{align}
    This corollary can be proved by applying Jensen's inequality to (\ref{lemma: martingale eq-0}), 
\end{corollary}
Notably, Corollary \ref{lemma: supermartingale} offers a useful insight that a supermartingale that involves the log-likelihood ratio could still be constructed despite that $\ell_{t}(\MLE{t}) - \ell_{t}(\btheta^{*})$ is a \textit{submartingale}. This result generalizes the classic result in \citep[Lemma 3]{kumar1982optimal} for tabular MDPs to the linear MDP setting, and is also holds for non-regularized MLE ($\lambda = 0$).
To establish Theorem \ref{lemma: central limit theorem}}, we define a useful quantity $\Delta_t$ as
\begin{equation}
    \Delta_t := \sum_{i=1}^{t-1} z_i = \sum_{i=1}^{t-1} \log\bigg({\frac{\phi_{i}(s_{i+1})^\top\MLE{t}}{\phi_{i}(s_{i+1})^\top\MLE{i}}}\bigg), \label{eq: Delta_t}
\end{equation}
where $\phi_i(s):= \phi(s|s_i,a_i)$ is a shorthand for the feature vector. In the following lemma, we present an upper bound for $\Delta_t$. Recall that $\lVert \phi(s'|s,a)\rVert_2 \leq L$, for all $s,s'\in \cS$ and $a\in \cA$.
\begin{lemma}
    \label{lemma: regret of online learning}
    For all $\lambda \geq 0$, we have
    \begin{align}
        \Delta_t \leq \frac{8d^2}{p^2_{\text{min}}}\log\left( \frac{d\lambda+(t-1)L^2}{d}\right). \label{lemma: regret of online learning eq-1}
    \end{align}
\end{lemma}

\begin{remark}[Connection between linear MDPs and online learning]
    \normalfont Through $\Delta_t$ and Lemma \ref{lemma: regret of online learning}, we could build an interesting connection between MLE in linear MDPs and the \textit{Follow-the-Leader} algorithm \citep{gaivoronski2000stochastic} in online learning. The connection is two-fold: (i) \textit{MLE in linear MDPs can be viewed as a variant of online portfolio selection problem}: We find that the MLE optimization problem in linear MDPs takes the same form as the classic online portfolio selection problem \citep{hazan2016introduction}. Specifically, the feature vectors and the dynamics model parameter in linear MDPs correspond to the price vectors and the asset allocation, respectively. The main difference of the two problems lies in the feasible set and the constraints. (ii) \textit{Iterative MLE is equivalent to Follow-the-Leader algorithm}: Another interesting connection is that applying MLE in each time step would correspond to the classic Follow-the-Leader algorithm \citep{gaivoronski2000stochastic}. Moreover, the term $\Delta_t$ in \ref{lemma: regret of online learning eq-1} could be interpreted as the regret of the Follow-the-Leader algorithm in online learning. With that said, one could verify that Lemma \ref{lemma: regret of online learning} is consistent with the regret quantified in \citep{gaivoronski2000stochastic}.
\end{remark}


Based on the supporting lemmas introduced above, we are ready to formally present the convergence result of MLE in linear MDPs.
\begin{theorem}
    \label{lemma: central limit theorem}
    With probability at least $1-\delta$, we have
\begin{align}
    \lVert \btheta^* - \MLE{t} \rVert^2_{\bA_t} \leq \CLT{t}, \label{lemma: central limit theorem eq-1}
\end{align}
\end{theorem}
The complete proof of Theorem \ref{lemma: central limit theorem} is provided in Appendix \ref{app:thm:theta_MLE}, and here we provide a proof sketch:

\textit{Proof Sketch.} Based on the result in Lemma \ref{lemma: supermartingale}, we can apply Azuma–Hoeffding inequality presented in Lemma \ref{lemma: Azuma–Hoeffding inequality} to get the high probability bound of log-likelihood ratio.
There are two main challenges that need to be handled: (i) The first one is the additional term $\Delta_t$, we find a connection to the analysis of the online portfolio selection problem and use a similar approach to handle it.
(ii) The other one is $M_t$, which represents the cumulative difference of the super-martingale. We adopt a similar approach by considering a stopping time to ensure that this theorem holds with high probability.
\subsection{Regret Bound of VBMLE}
In this subsection, we formally provide the regret bound of the proposed VBMLE algorithm.
\begin{theorem}
\label{theorem: upper bound of regret}
    For all linear kernel MDP $\mathcal{M} = \langle \mathcal{S}, \mathcal{A}, P, R, T, \mu_0\rangle$, with probability at least $1-\frac{1}{T} - 3\delta$ and choosing $\alpha(t) = \sqrt{t}$, VBMLE, proposed in Algorithm \ref{alg:VBMLE}, has a regret upper bound that satisfies 
    \begin{align}
        \mathcal{R}(T) = \mathcal{O}\left(\frac{d\sqrt{T}\log{T}}{(1-\gamma)^2}\right)\label{theorem: upper bound of regret eq-0}. 
    \end{align}
\end{theorem}
The complete proof of Theorem \ref{theorem: upper bound of regret} is provided in Appendix \ref{app:thm:regret}, and here we provide a proof sketch:

\textit{Proof Sketch.} 
\begin{enumerate}[leftmargin=*]
    \item Similar to the analysis of the upper-confidence bound approach, which uses the concentration inequality to replace the term associated with an optimal policy, under VBMLE we can replace $V^*(s_t,\btheta^*)$ by applying the objective function of VBMLE. 
    \item Then, there are two terms that need to be handled: (i) $\lVert \MLE{t} - \btheta^* \rVert_{\bA_t}$ and (ii) $\lVert \bthetaR - \btheta^* \rVert_{\bA_t}$. We provide a novel theorem of the confidence ellipsoid of the maximum likelihood estimator in linear MDPs in Theorem \ref{lemma: central limit theorem} to deal with (i).
    \item In contrast to the regret analysis presented in \citep{hung2021reward}, where the likelihood of an exponential family distribution was considered, analyzing regret in the linear MDP setting is more complex due to the absence of simple closed-form expressions for both $\MLE{t}$ and $\bthetaR$.  Additionally, in this context, the bias term is not linear with respect to $\btheta$, even if we represent it as $Q^*(s_t,a_t;\btheta) = \innerprod{\sum_{s'\in\mathcal{S}}\phi(s'|s_t,a_t)V^*(s_t,\btheta)}{\btheta}$. To address these challenges, we adopt a novel approach by completing the square of $\lVert \bthetaR - \btheta^* \rVert_{\bA_t}$ and successfully overcome the problems mentioned above.
\end{enumerate}

\section{Numerical Experiments}
We demonstrate the empirical performance of VBMLE in terms of both regret and computation time in this section. we conduct experiments on a simple environment with discrete state and action spaces. To provide a detailed understanding of how we transitioned from the tabular MDP to the linear MDP setting, we have outlined the procedure in the Appendix \ref{section: env}. The following result includes a comparison between VBMLE and UCLK, a well-known algorithm used in the context of infinite horizon linear MDPs. Details regarding the selected hyperparameters can be found in Appendix \ref{section: hyperparameter}.

\begin{figure}[!ht]
\centering
$\begin{array}{c c}
    \multicolumn{1}{l}{\mbox{\bf }} & \multicolumn{1}{l}{\mbox{\bf }} \\
    \scalebox{0.49}{\includegraphics[width=\textwidth]{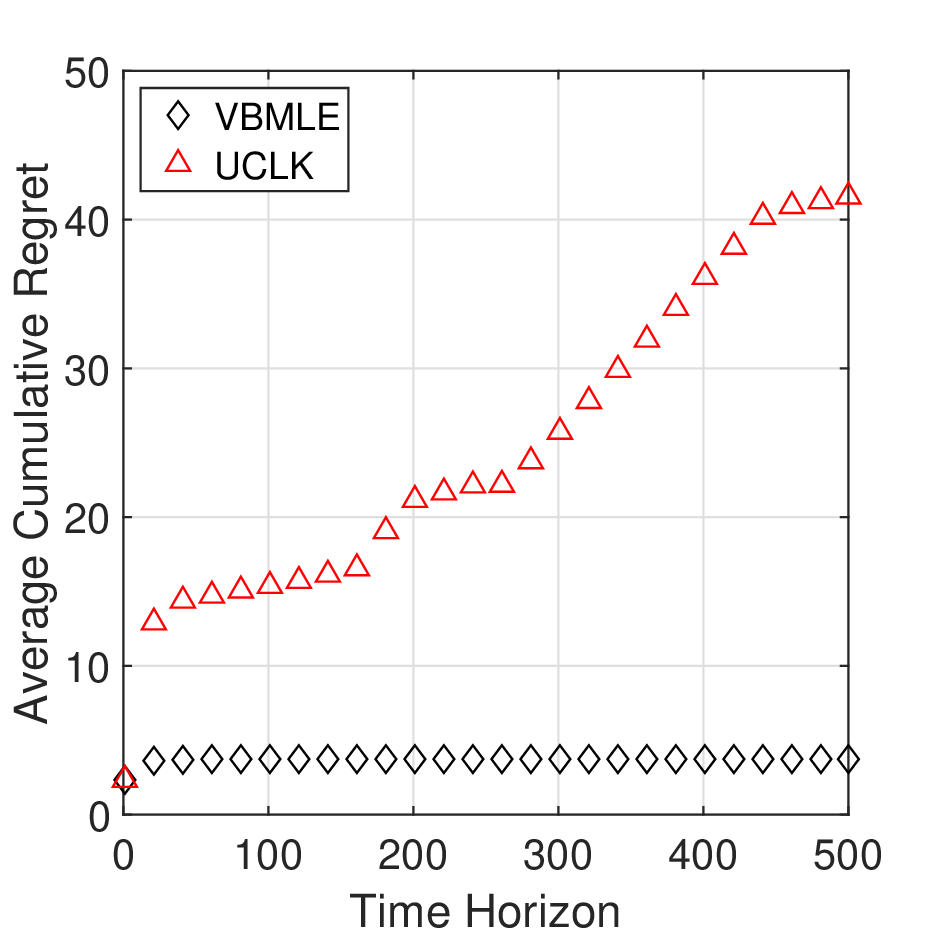}} &  \scalebox{0.49}{\includegraphics[width=\textwidth]{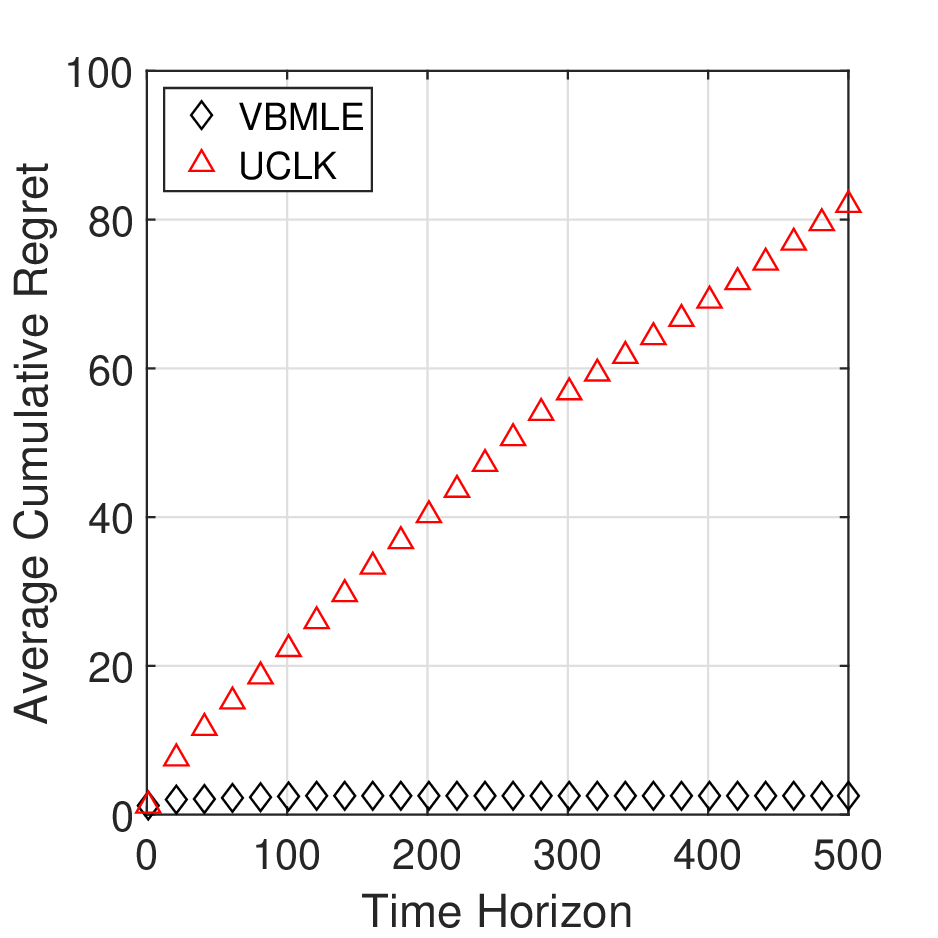}} \\
    \mbox{(a) $|\mathcal{S}| = 3, |\mathcal{A}| = 2$} &  \mbox{(b) $|\mathcal{S}| = 5, |\mathcal{A}| = 4$}
\end{array}$
\caption{Regret averaged over $20$ trials with $T = 500$.}
\label{fig:regret}
\end{figure}

\begin{figure}[!ht]
\centering
$\begin{array}{c c}
    \multicolumn{1}{l}{\mbox{\bf }} & \multicolumn{1}{l}{\mbox{\bf }} \\
    \scalebox{0.49}{\includegraphics[width=\textwidth]{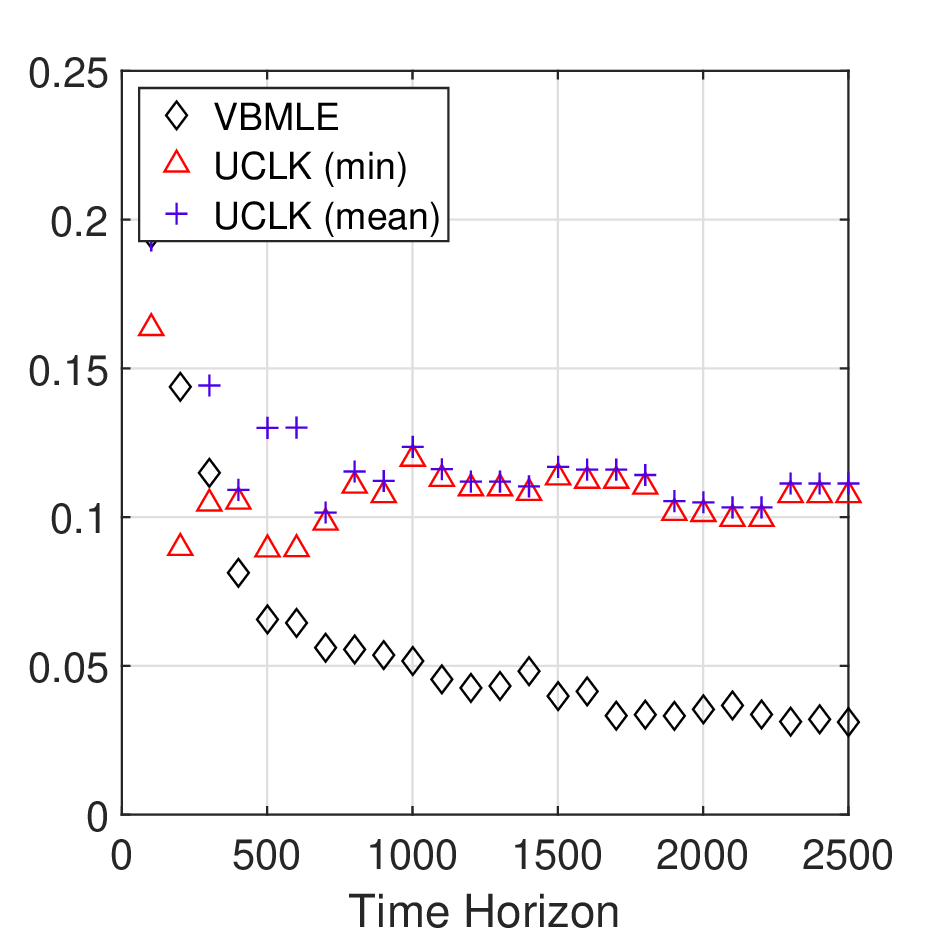}} &  \scalebox{0.49}{\includegraphics[width=\textwidth]{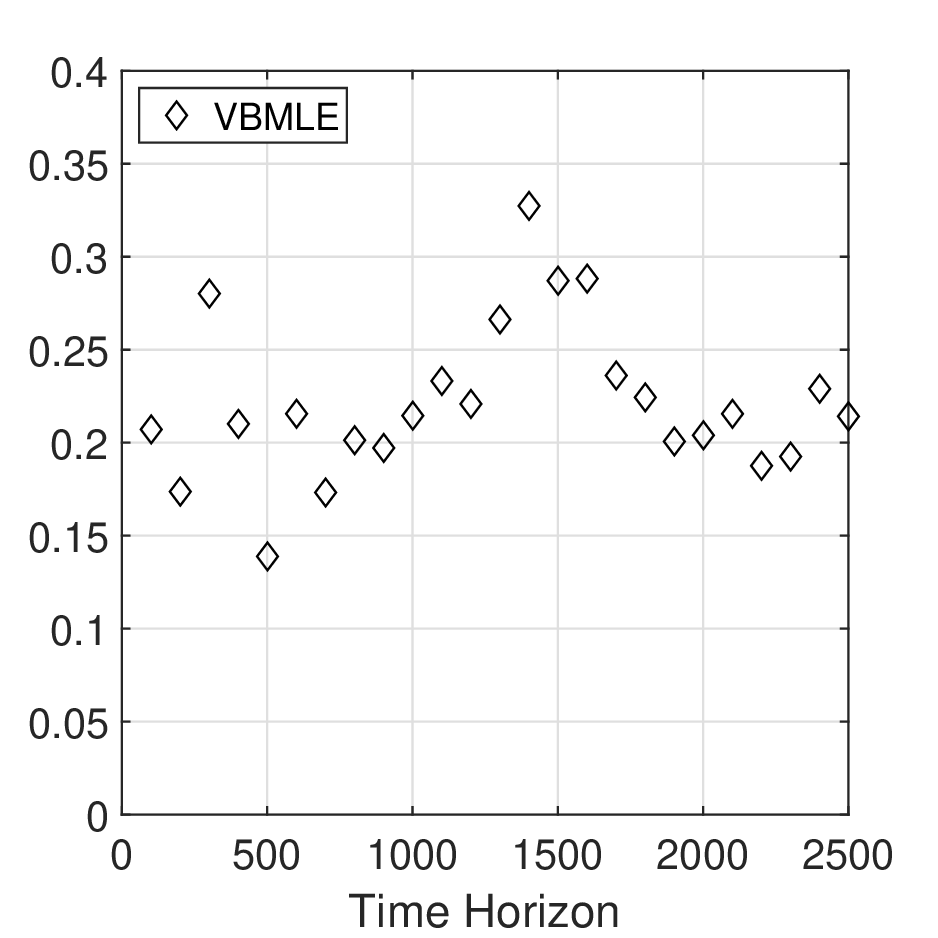}} \\
    \mbox{(a) Distance between learned $\theta$ and $\theta^*$} &  \mbox{(b) $\lVert\MLE{t} - \btheta^*\rVert^2_{\bA_t}$}
\end{array}$
\caption{Observation of the distance over with $T = 2500$.}
\label{fig:theta}
\end{figure}

\begin{itemize}[leftmargin=*]
    \item {\bf Empirical Regret:} Figure \ref{fig:regret} provides the empirical regret of VBMLE and UCLK across various sizes of linear MDPs, and the results demonstrate that VBMLE outperforms UCLK in terms of regret performance. We also provide the standard deviation of the regret at the final step in Table \ref{table: std of regret}. \hungyh{VBMLE also has better robustness with an order of magnitude smaller standard deviation than UCLK.}
    \item {\bf Computation Time:} \hungyh{Theoratically, UCLK requires $U|\mathcal{S}||\mathcal{A}|$ times of solving the optimization problem per step, where $U$ is the times of value iteration, and VBMLE only requires once.} Table \ref{table: computation time} displays the computation time per step within the same environment as depicted in Figure \ref{fig:regret}(a). It is evident that the computational complexity of UCLK renders the algorithm impractical in large MDP settings.
    \item {\bf Distance between the learned $\btheta$ and $\btheta^*$:} Figure \ref{fig:theta}(a) shows the comparison of $\lVert\bthetaR - \btheta^*\rVert^2_{2}$ and $\lVert\btheta^{\textbf{UCLK}}_t - \btheta^*\rVert^2_{2}$. Notably, due to UCLK learning distinct parameter for each state-action pair, we also plot  $\min_{s,a}\lVert\btheta^{\textbf{UCLK}}_t(s,a) - \btheta^*\rVert^2_{2}$ and $\frac{1}{|\mathcal{S}||\mathcal{A}|}\sum_{(s,a)}\lVert\btheta^{\textbf{UCLK}}_t(s,a) - \btheta^*\rVert^2_{2}$ for UCLK. The result shows that VBMLE learned a more accurate representation of true parameter $\btheta^*$.
\end{itemize}

\begin{table}[!ht]
\label{table: computation time}
\centering
\caption{Computation time per step of VBMLE and UCLK under different sizes of tabular MDPs.}

\begin{tabular}{|c|c|c|c|}
\hline
               & $|\mathcal{S}| = 3, |\mathcal{A}| = 2$ & $|\mathcal{S}| = 5, |\mathcal{A}| = 4$ & $|\mathcal{S}| = 15, |\mathcal{A}| = 4$ \\ \hline
\textbf{UCLK}  & $3.135$s                                & $49.763$s                               & $\geq 35\text{hr}$                      \\ \hline
\textbf{VBMLE} & $0.793$s                                & $2.359$s                                & $42.232$s                                \\ \hline
\end{tabular}
\end{table}
\section{Conclusion}
We proposed a provably effective and computationally efficient algorithm for solving linear MDPs, called VBMLE. The sample complexity of the proposed is proved to be upper bounded by $\mathcal{O}\left(d\sqrt{T}\log{T}/(1-\gamma)^2\right)$. The proposed algorithm is different from the traditional value-target regression approach and leverages the MLE with value bias to learn the dynamic. We provide a novel theorem to show the confidence ellipsoid of MLE and the simulation result demonstrates the empirical performance of VBMLE.

\bibliography{iclr2024_conference}

\begin{thebibliography}{24}
\providecommand{\natexlab}[1]{#1}
\providecommand{\url}[1]{\texttt{#1}}
\expandafter\ifx\csname urlstyle\endcsname\relax
  \providecommand{\doi}[1]{doi: #1}\else
  \providecommand{\doi}{doi: \begingroup \urlstyle{rm}\Url}\fi

\bibitem[Abbasi-Yadkori et~al.(2011)Abbasi-Yadkori, P{\'a}l, and Szepesv{\'a}ri]{abbasi2011improved}
Yasin Abbasi-Yadkori, D{\'a}vid P{\'a}l, and Csaba Szepesv{\'a}ri.
\newblock Improved algorithms for linear stochastic bandits.
\newblock \emph{Advances in neural information processing systems}, 24:\penalty0 2312--2320, 2011.

\bibitem[Auer et~al.(2008)Auer, Jaksch, and Ortner]{auer2008near}
Peter Auer, Thomas Jaksch, and Ronald Ortner.
\newblock Near-optimal regret bounds for reinforcement learning.
\newblock \emph{Advances in neural information processing systems}, 21, 2008.

\bibitem[Ayoub et~al.(2020)Ayoub, Jia, Szepesvari, Wang, and Yang]{ayoub2020model}
Alex Ayoub, Zeyu Jia, Csaba Szepesvari, Mengdi Wang, and Lin Yang.
\newblock Model-based reinforcement learning with value-targeted regression.
\newblock In \emph{International Conference on Machine Learning}, pp.\  463--474. PMLR, 2020.

\bibitem[Azar et~al.(2017)Azar, Osband, and Munos]{azar2017minimax}
Mohammad~Gheshlaghi Azar, Ian Osband, and R{\'e}mi Munos.
\newblock Minimax regret bounds for reinforcement learning.
\newblock In \emph{International Conference on Machine Learning}, pp.\  263--272. PMLR, 2017.

\bibitem[Borkar \& Varaiya(1979)Borkar and Varaiya]{borkar1979adaptive}
Vivek Borkar and P~Varaiya.
\newblock Adaptive control of {M}arkov chains, {I}: {F}inite parameter set.
\newblock \emph{IEEE Transactions on Automatic Control}, 24\penalty0 (6):\penalty0 953--957, 1979.

\bibitem[Cai et~al.(2020)Cai, Yang, Jin, and Wang]{cai2020provably}
Qi~Cai, Zhuoran Yang, Chi Jin, and Zhaoran Wang.
\newblock Provably efficient exploration in policy optimization.
\newblock In \emph{International Conference on Machine Learning}, pp.\  1283--1294. PMLR, 2020.

\bibitem[Chen et~al.(2022)Chen, He, and Gu]{chen2022sample}
Yuanzhou Chen, Jiafan He, and Quanquan Gu.
\newblock On the sample complexity of learning infinite-horizon discounted linear kernel mdps.
\newblock In \emph{International Conference on Machine Learning}, pp.\  3149--3183. PMLR, 2022.

\bibitem[Gaivoronski \& Stella(2000)Gaivoronski and Stella]{gaivoronski2000stochastic}
Alexei~A Gaivoronski and Fabio Stella.
\newblock Stochastic nonstationary optimization for finding universal portfolios.
\newblock \emph{Annals of Operations Research}, 100:\penalty0 165--188, 2000.

\bibitem[Hazan et~al.(2016)]{hazan2016introduction}
Elad Hazan et~al.
\newblock Introduction to online convex optimization.
\newblock \emph{Foundations and Trends{\textregistered} in Optimization}, 2\penalty0 (3-4):\penalty0 157--325, 2016.

\bibitem[Hung \& Hsieh(2023)Hung and Hsieh]{hung2023reward}
Yu-Heng Hung and Ping-Chun Hsieh.
\newblock {Reward-biased maximum likelihood estimation for neural contextual bandits: A distributional learning perspective}.
\newblock In \emph{Proceedings of the AAAI Conference on Artificial Intelligence}, volume~37, pp.\  7944--7952, 2023.

\bibitem[Hung et~al.(2021)Hung, Hsieh, Liu, and Kumar]{hung2021reward}
Yu-Heng Hung, Ping-Chun Hsieh, Xi~Liu, and P.~R. Kumar.
\newblock Reward-biased maximum likelihood estimation for linear stochastic bandits.
\newblock In \emph{Proceedings of the AAAI Conference on Artificial Intelligence}, pp.\  7874--7882, 2021.

\bibitem[Jin et~al.(2020)Jin, Yang, Wang, and Jordan]{jin2020provably}
Chi Jin, Zhuoran Yang, Zhaoran Wang, and Michael~I Jordan.
\newblock Provably efficient reinforcement learning with linear function approximation.
\newblock In \emph{Conference on Learning Theory}, pp.\  2137--2143. PMLR, 2020.

\bibitem[Kumar \& Lin(1982)Kumar and Lin]{kumar1982optimal}
P~Kumar and Woei Lin.
\newblock Optimal adaptive controllers for unknown markov chains.
\newblock \emph{IEEE Transactions on Automatic Control}, 27\penalty0 (4):\penalty0 765--774, 1982.

\bibitem[Kumar \& Varaiya(2015)Kumar and Varaiya]{kumar2015stochastic}
PR~Kumar and Pravin Varaiya.
\newblock \emph{{Stochastic Systems: Estimation, Identification, and Adaptive Control}}, volume~75.
\newblock SIAM, 2015.

\bibitem[Liu et~al.(2020)Liu, Hsieh, Hung, Bhattacharya, and Kumar]{liu2020exploration}
Xi~Liu, Ping-Chun Hsieh, Yu~Heng Hung, Anirban Bhattacharya, and P.~R. Kumar.
\newblock {Exploration Through Reward Biasing: Reward-Biased Maximum Likelihood Estimation for Stochastic Multi-Armed Bandits}.
\newblock In \emph{International Conference on Machine Learning}, pp.\  6248--6258. PMLR, 2020.

\bibitem[Mete et~al.(2021)Mete, Singh, Liu, and Kumar]{mete2021reward}
Akshay Mete, Rahul Singh, Xi~Liu, and P.~R. Kumar.
\newblock Reward biased maximum likelihood estimation for reinforcement learning.
\newblock In \emph{Learning for Dynamics and Control}, pp.\  815--827, 2021.

\bibitem[Mete et~al.(2022)Mete, Singh, and Kumar]{mete2022augmented}
Akshay Mete, Rahul Singh, and P.~R. Kumar.
\newblock {Augmented RBMLE-UCB Approach for Adaptive Control of Linear Quadratic Systems}.
\newblock In \emph{Advances in Neural Information Processing Systems}, 2022.

\bibitem[Modi et~al.(2020)Modi, Jiang, Tewari, and Singh]{modi2020sample}
Aditya Modi, Nan Jiang, Ambuj Tewari, and Satinder Singh.
\newblock Sample complexity of reinforcement learning using linearly combined model ensembles.
\newblock In \emph{International Conference on Artificial Intelligence and Statistics}, pp.\  2010--2020. PMLR, 2020.

\bibitem[Wang et~al.(2020)Wang, Salakhutdinov, and Yang]{wang2020reinforcement}
Ruosong Wang, Russ~R Salakhutdinov, and Lin Yang.
\newblock Reinforcement learning with general value function approximation: Provably efficient approach via bounded eluder dimension.
\newblock \emph{Advances in Neural Information Processing Systems}, 33:\penalty0 6123--6135, 2020.

\bibitem[Wang et~al.(2019)Wang, Wang, Du, and Krishnamurthy]{wang2019optimism}
Yining Wang, Ruosong Wang, Simon~S Du, and Akshay Krishnamurthy.
\newblock Optimism in reinforcement learning with generalized linear function approximation.
\newblock \emph{arXiv preprint arXiv:1912.04136}, 2019.

\bibitem[Yang \& Wang(2019)Yang and Wang]{yang2019sample}
Lin Yang and Mengdi Wang.
\newblock Sample-optimal parametric q-learning using linearly additive features.
\newblock In \emph{International Conference on Machine Learning}, pp.\  6995--7004. PMLR, 2019.

\bibitem[Zanette et~al.(2020)Zanette, Lazaric, Kochenderfer, and Brunskill]{zanette2020learning}
Andrea Zanette, Alessandro Lazaric, Mykel Kochenderfer, and Emma Brunskill.
\newblock Learning near optimal policies with low inherent bellman error.
\newblock In \emph{International Conference on Machine Learning}, pp.\  10978--10989. PMLR, 2020.

\bibitem[Zhou et~al.(2021{\natexlab{a}})Zhou, Gu, and Szepesvari]{zhou2021nearly}
Dongruo Zhou, Quanquan Gu, and Csaba Szepesvari.
\newblock Nearly minimax optimal reinforcement learning for linear mixture markov decision processes.
\newblock In \emph{Conference on Learning Theory}, pp.\  4532--4576. PMLR, 2021{\natexlab{a}}.

\bibitem[Zhou et~al.(2021{\natexlab{b}})Zhou, He, and Gu]{zhou2021provably}
Dongruo Zhou, Jiafan He, and Quanquan Gu.
\newblock Provably efficient reinforcement learning for discounted mdps with feature mapping.
\newblock In \emph{International Conference on Machine Learning}, pp.\  12793--12802. PMLR, 2021{\natexlab{b}}.

\end{thebibliography}
\bibliographystyle{iclr2024_conference}

\newpage
\appendix
\section{Proofs of the Supporting Lemmas}
To begin with, we provide several useful definitions as follows.
Define the likelihood ratio as
\begin{equation}
  L_t(\btheta) := \prod_{i=1}^{t-1} \frac{\text{Pr}(s_{i+1}|s_i,a_i;\btheta)}{\text{Pr}(s_{i+1}|s_i,a_i;\btheta^*)} \cdot \text{exp}\left( \frac{\lambda}{2}\lVert \btheta \rVert_2^2\right). \label{eq: likelihood ratio}
\end{equation}

\begin{lemma}
    \label{lemma: submartingale}
    \begin{align}
        L_t(\MLE{t}) := \max_{\btheta \in \mathbb{P}}L_t(\btheta)
    \end{align}
    is a sub-martingale.
    \begin{proof}
        Given the causal information $\mathcal{F}_t$, we have
        \begin{align}
            \mathbb{E}_{s_{t+1}\sim \text{Pr}(\cdot|s_t,a_t;\btheta^{*})}\left[ \max_{\btheta\in\mathbb{P}}L_{t+1}(\btheta)\big\rvert\mathcal{F}_t \right] \geq & \max_{\btheta\in\mathbb{P}} \mathbb{E}_{s_{t+1}\sim \text{Pr}(\cdot|s_t,a_t;\btheta^{*})}\left[ L_{t+1}(\btheta)\big\rvert\mathcal{F}_t \right] \label{lemma: submartingale eq-1}\\
            = &\max_{\btheta\in\mathbb{P}} \left\{\sum_{s'\in\mathcal{S}} \text{Pr}(s'|s_t,a_t;\btheta^*)\cdot\frac{\text{Pr}(s'|s_t,a_t;\btheta)}{\text{Pr}(s'|s_t,a_t;\btheta^*)} \cdot L_t(\btheta)\right\} \label{lemma: submartingale eq-2}\\
            = & L_t(\MLE{t}) \label{lemma: submartingale eq-3},
        \end{align}
        where (\ref{lemma: submartingale eq-1}) holds by $\max\limits_{\btheta\in\mathbb{P}}L_{t+1}(\btheta) \geq L_{t+1}(\btheta')$, and 
        \begin{align}
            \btheta' := \argmax\limits_{\btheta\in\mathbb{P}} \left\{\mathbb{E}_{s_{t+1}\sim \text{Pr}(\cdot|s_t,a_t;\btheta^{*})}\left[ L_{t+1}(\btheta)\big\rvert\mathcal{F}_t \right]\right\} \label{lemma: submartingale eq-4}.
        \end{align}
    \end{proof}
\end{lemma}
\begin{remark}
    \normalfont
    The lemma shows that it is not able to apply Azuma–Hoeffding inequality on the maximum likelihood ratio. However, we can still add an additional term to construct a supermartingale:
    \begin{align}
        \mathbb{E}_{s_{t+1}\sim \text{Pr}(\cdot|s_t,a_t;\btheta^{*})}\left[ \max_{\btheta\in\mathbb{P}}L_{t+1}(\btheta) \cdot \prod_{i=1}^{t}\left( \sum_{s'\in \mathcal{S}} h_t(s')\right)^{-1}\big\rvert\mathcal{F}_t \right] = L_t(\MLE{t})\cdot \prod_{i=1}^{t-1}\left( \sum_{s'\in \mathcal{S}} h_t(s')\right)^{-1}
    \end{align}
    where
    \begin{align}
        h_t(s'):=\max_{\btheta\in\mathbb{P}}\left\{\frac{L_t(\btheta)}{L_t(\MLE{t})}\cdot \text{Pr}(s'|s_t,a_t;\btheta)\right\}.
    \end{align}
    It's important to note that there are two primary challenges when dealing with $h_t(s')$:
    \begin{itemize}
        \item Dependency on state space size: The first challenge arises from the fact that $\sum_{s'\in\mathcal{S}} h_t(s')$ exhibits a dependency on the size of the state space. This dependence can complicate the regret analysis, especially in cases where the state space is large.
        \item State-dependent maximization: The second challenge is related to the maximization inside $h_t(s')$, which is also influenced by the specific state $s'$ under consideration. Formally, the absence of a closed-form expression for this maximization makes it challenging to conduct a straightforward analysis.
    \end{itemize}
    To address the above two challenges, we then introduce a new supermartingale associate to the likelihood ratio evaluated on the previous step's maximum likelihood estimator.
\end{remark}

\subsection{Proof of Lemma \ref{lemma: supermartingale}}

Recall that
\begin{align}
      X_t &:= \ell_{t}(\MLE{t-1}) - \ell_{t}(\btheta^{*}) + \sum_{i=1}^{t-1} z_i,\\
      z_t &:= \ell_{t}(\MLE{t}) - \ell_{t}(\MLE{t-1}).
\end{align}
For ease of exposition, we restate Lemma \ref{lemma: supermartingale} as follows.
\hungyh{
\begin{lemma*}
  The stochastic process $\{L_{t}(\MLE{t-1}) \cdot \prod_{i=1}^{t-1}\text{\normalfont exp}(-z_i)\}$ is a martingale, i.e.,
    \begin{align}
        \normalfont
        \mathbb{E}_{s_{t+1}\sim \text{Pr}(\cdot|s_t,a_t;\btheta^{*})}\left[ L_{t+1}(\MLE{t})\cdot\prod_{i=1}^{t}\text{exp}(-z_i)\bigg\rvert\mathcal{F}_t \right] = L_{t}(\MLE{t-1}) \cdot \prod_{i=1}^{t-1}\text{exp}(-z_i), 
    \end{align}
    where $\mathcal{F}_{t}:= \{s_1,a_1,\cdots,s_{t},a_{t}\}$ denotes the causal information up to time $t$.  
\end{lemma*}
}
\begin{proof}
     Recall the definition of likelihood ratio in (\ref{eq: likelihood ratio}). We have
    \begin{align}
        &\mathbb{E}_{s_{t+1}\sim \text{Pr}(\cdot|s_t,a_t;\btheta^{*})}\left[ L_{t+1}(\MLE{t})\cdot\prod_{i=1}^{t}\text{exp}(-z_i)\bigg\rvert\mathcal{F}_t \right] \nonumber\\
        =& \sum_{s'\in \mathcal{S}} \text{Pr}(s'|s_t,a_t;\btheta^*)\cdot L_{t}(\MLE{t}) \cdot \frac{\text{Pr}(s'|s_t,a_t;\MLE{t})}{\text{Pr}(s'|s_t,a_t;\btheta^*)} \cdot \prod_{i=1}^{t}\text{exp}(-z_i), \label{lemma: supermartingale eq-1}\\
        = & L_t(\MLE{t})\cdot \prod_{i=1}^{t}\text{exp}(-z_i), \label{lemma: supermartingale eq-2}\\
        = & L_{t}(\MLE{t-1}) \cdot \prod_{i=1}^{t-1}\text{exp}(-z_i)\label{lemma: supermartingale eq-3},
    \end{align}
    where (\ref{lemma: supermartingale eq-1}) holds by the definition of the expectation, (\ref{lemma: supermartingale eq-2})  holds due to $\MLE{t} \in \mathbb{P}$ and $\mathcal{F}_t$-measurability, and (\ref{lemma: supermartingale eq-3}) holds by $\exp(z_t) = \frac{L_t(\MLE{t})}{L_t(\MLE{t-1})}$. \hungyh{We complete this proof. }
\end{proof}

\subsection{Proof of Lemma \ref{lemma: regret of online learning}}

Recall the definition of $\Delta_t$ from (\ref{eq: Delta_t}) as
\begin{equation}
    \Delta_t := \sum_{i=1}^{t-1} \log{\bigg(\frac{\phi_{i}(s_{i+1})^\top\MLE{t}}{\phi_{i}(s_{i+1})^\top\MLE{i}}\bigg)} + \frac{\lambda}{2} \left(\lVert \MLE{t} \rVert_2^2 - \lVert \MLE{0} \rVert_2^2\right),
\end{equation}
where $\phi_i(s):= \phi(s|s_i,a_i)$ is a shorthand for the feature vector. 
For ease of exposition, we restate Lemma \ref{lemma: regret of online learning} as follows.

\begin{lemma*}
    For all $\lambda \geq 0$, we have that
    \begin{align}
        \Delta_t \leq \frac{8d^2}{p^2_{\text{min}}}\log\left( \frac{d\lambda+(t-1)L^2}{d}\right). 
    \end{align}
\end{lemma*}
\begin{proof}
To begin with, we have 
        \begin{align}
            \Delta_{t+1} =& \sum_{i=1}^{t-1} \log{\phi_{i}(s_{i+1})^\top\MLE{t+1}} - \sum_{i=1}^{t-1} \log{\phi_{i}(s_{i+1})^\top\MLE{i}}  \nonumber \\
            & + \frac{\lambda}{2} \left(\lVert \MLE{t+1} \rVert_2^2 - \lVert \MLE{0} \rVert_2^2\right)+ \log{\frac{\phi_{t}(s_{t+1})^\top\MLE{t+1}}{\phi_{t}(s_{t+1})^\top\MLE{t}}}  \label{lemma: regret of online learning eq-3} \\
            \leq & \sum_{i=1}^{t-1} \log{\phi_{i}(s_{i+1})^\top\MLE{t}} - \sum_{i=1}^{t-1} \log{\phi_{i}(s_{i+1})^\top\MLE{i}} \nonumber \\
            &  + \frac{\lambda}{2} \left(\lVert \MLE{t} \rVert_2^2 - \lVert \MLE{0} \rVert_2^2\right)  + \log{\frac{\phi_{t}(s_{t+1})^\top\MLE{t+1}}{\phi_{t}(s_{t+1})^\top\MLE{t}}}\label{lemma: regret of online learning eq-4} \\
            \leq & \sum_{i=1}^{t-1} \log{\phi_{i}(s_{i+1})^\top\MLE{t}} - \sum_{i=1}^{t-1} \log{\phi_{i}(s_{i+1})^\top\MLE{i}} \nonumber \\
            &  + \frac{\lambda}{2} \left(\lVert \MLE{t} \rVert_2^2 - \lVert \MLE{0} \rVert_2^2\right)  + \frac{\innerprod{\phi_t(s_{t+1})}{\MLE{t+1} - \MLE{t}}}{\phi_{t}(s_{t+1})^\top\MLE{t}}\label{lemma: regret of online learning eq-4.5} \\
            \leq & \Delta_t + \Big\lVert \frac{\phi_t(s_{t+1})}{\phi_{t}(s_{t+1})^\top\MLE{t}} \Big\rVert_{\bA^{-1}} \cdot \lVert \MLE{t+1} - \MLE{t} \rVert_{\bA} \label{lemma: regret of online learning eq-5}, 
        \end{align}
        where (\ref{lemma: regret of online learning eq-4}) holds by $\MLE{t} = \argmax_{\btheta\in\mathbb{P}}\ell_t(\btheta)$, (\ref{lemma: regret of online learning eq-4.5}) holds by the fact that $\log(x+1)\leq x$, and (\ref{lemma: regret of online learning eq-5}) holds by Cauchy–Schwarz inequality under any positive definite matrix $\bA$.  To handle $\lVert \MLE{t+1} - \MLE{t} \rVert_{\bA}$, we start from the fact that $\MLE{t+1} = \argmax_{\btheta\in\mathbb{P}}\ell_{t+1}(\btheta)$, which leads to the following inequality:
        \begin{align}
            0 \leq & \ell_{t+1}(\MLE{t+1}) - \ell_{t+1}(\MLE{t}) \label{lemma: regret of online learning eq-6} \\
            = & (\MLE{t+1} - \MLE{t})^\top\nabla_{\btheta}\left(\ell_t(\btheta) + \log\phi_t(s_{t+1})^\top\btheta\right)\big\rvert_{\btheta = \MLE{t}} -\frac{1}{2}\lVert \MLE{t+1} - \MLE{t}\rVert^2_{\bA_{t+1}(\btheta'_{t+1})} \label{lemma: regret of online learning eq-7} \\
            \leq & \frac{\phi_t(s_{t+1})^\top}{\phi_t(s_{t+1})^\top\MLE{t}}(\MLE{t+1} - \MLE{t}) -\frac{1}{2}\lVert \MLE{t+1} - \MLE{t}\rVert^2_{\bA_{t+1}(\btheta'_{t+1})} \label{lemma: regret of online learning eq-8},
        \end{align}
        \pch{where (\ref{lemma: regret of online learning eq-7}) holds by Taylor's theorem and $\bA_{t+1}(\btheta'_{t+1}):= -\nabla^2_{\btheta}\ell_{t+1}(\btheta)|_{\btheta = \btheta'_{t+1}}$ ($\btheta'_{t+1}$ is some convex combination between $\MLE{t+1}$ and $\MLE{t}$), and (\ref{lemma: regret of online learning eq-8}) holds due to the necessary condition of optimality for constrained problem as $\nabla_{\btheta}\ell_t(\btheta)|_{\btheta = \MLE{t}}^\top(\MLE{t+1} - \MLE{t}) \leq 0$.} Then, by reordering (\ref{lemma: regret of online learning eq-8}) and applying Cauchy–Schwarz inequality, we have 
        \begin{align}
            \frac{1}{2}\lVert \MLE{t+1} - \MLE{t}\rVert_{\bA_{t+1}(\btheta'_{t+1})} \leq \lVert \frac{\phi_t(s_{t+1})^\top}{\phi_t(s_{t+1})^\top\MLE{t}} \rVert_{\bA^{-1}_{t+1}(\btheta'_{t+1})} \label{lemma: regret of online learning eq-9}.
        \end{align}
        By plugging $\bA=\bA_{t+1}(\btheta'_{t+1})$ into (\ref{lemma: regret of online learning eq-5}) as well as combining (\ref{lemma: regret of online learning eq-5}) and (\ref{lemma: regret of online learning eq-9}), we have
        \begin{align}
            \Delta_{t+1} \leq & \Delta_{t} + 2 \Big\lVert \frac{\phi_t(s_{t+1})}{\phi_t(s_{t+1})^\top\MLE{t}} \Big\rVert^2_{\bA^{-1}_{t+1}(\btheta'_{t+1})}, \label{lemma: regret of online learning eq-10}\\
            \leq &\Delta_{t} + \frac{2}{p^2_{\text{min}}}\cdot \lVert \phi_t(s_{t+1}) \rVert^2_{\bA^{-1}_{t}}, \label{lemma: regret of online learning eq-11}
        \end{align}
        where (\ref{lemma: regret of online learning eq-11}) holds by Assumption \ref{assumption: p_min} and $\bA^{-1}_{t+1}(\btheta'_{t+1}) \preceq \bA^{-1}_{t+1} \preceq \bA^{-1}_{t}$ given the definition in (\ref{eq:A_t}). Then, by applying Lemma \ref{lemma: summation of vector norm} to (\ref{lemma: regret of online learning eq-11}), we have 
        \begin{align}
            \Delta_{t+1} \leq \frac{8d^2}{p^2_{\text{min}}}\log\left( \frac{d\lambda+ t L^2}{d}\right), \quad\forall t\in \mathbb{N}.
        \end{align}
\end{proof}

\section{Proofs of the Main Theorems}

\subsection{Proof of Theorem \ref{lemma: central limit theorem}}
\label{app:thm:theta_MLE}

For ease of exposition, we restate Theorem \ref{lemma: central limit theorem} as follows.
\begin{theorem*}
    At each time $t$, with probability at least $1-\delta$, we have
\begin{align}
    \lVert \btheta^* - \MLE{t} \rVert^2_{\bA_t} \leq \beta_t,
\end{align}
where $\beta_t := \frac{37d^2}{p^2_{\text{min}}}\cdot \log\left( \frac{d\lambda+tL^2}{d}\right)\cdot \log{\frac{1}{\delta}}$.
\end{theorem*}
\begin{proof}
By Corollary \ref{lemma: supermartingale} and Azuma–Hoeffding inequality in Lemma \ref{lemma: Azuma–Hoeffding inequality}, we have 
\begin{align}
    \text{Pr}\left(\ell_{t+1}(\MLE{t}) - \ell_{t+1}(\btheta^{*}) - \sum_{i=1}^{t}z_i \geq \sqrt{2M_t\log{\frac{1}{\delta}}}\right) \leq \delta, \label{lemma: central limit theorem eq-2}
\end{align}
where $M_t := \sum_{i=1}^{t} \left(\log{L_{i+1}(\MLE{i})} - \log{L_{i}(\MLE{i-1})} - z_i\right)^2$.
\begin{itemize}[leftmargin=*]
    \item Regarding $\ell_{t+1}(\MLE{t}) - \ell_{t+1}(\btheta^{*})$, we have
    \begin{align}
        \ell_{t+1}(\MLE{t}) - \ell_{t+1}(\btheta^{*}) = & \;\ell_{t}(\MLE{t}) - \ell_{t}(\btheta^{*}) +\log{\frac{\phi(s_{t+1}|s_t,a_t)^\top\MLE{t}}{\phi(s_{t+1}|s_t,a_t)^\top\btheta^*}} \label{lemma: central limit theorem eq-3} \\
        \geq & \;\frac{1}{2} \lVert \MLE{t} - \btheta^*\rVert^2_{-\nabla^2_{\btheta}\ell_{t}(\btheta)|_{\btheta = \btheta'}} +\log{\frac{\phi(s_{t+1}|s_t,a_t)^\top\MLE{t}}{\phi(s_{t+1}|s_t,a_t)^\top\btheta^*}} \label{lemma: central limit theorem eq-4} \\
        \geq & \;\frac{1}{2} \lVert \MLE{t} - \btheta^*\rVert^2_{\bA_t} - \frac{1}{p_{\text{min}}}, \label{lemma: central limit theorem eq-5}
    \end{align}
    where (\ref{lemma: central limit theorem eq-4}) holds by Taylor's theorem, the necessary condition of optimality for constrained problems $\nabla_{\btheta}\ell_t(\btheta)|_{\btheta= \MLE{t}}^\top(\btheta^* - \MLE{t})\leq 0$, and $\btheta'$ is some convex combination of $\MLE{t}$ and $\btheta^*$, and (\ref{lemma: central limit theorem eq-5}) holds by $\bA_t  \preceq -\nabla^2_{\btheta}\ell_{t}(\btheta)|_{\btheta = \btheta'}$, $\btheta^* \in \mathbb{P}$ and Assumption \ref{assumption: p_min}.
    \item For $\sum_{i=1}^{t}z_i$, denoting $\phi_t(s) := \phi(s|s_{t},a_{t})$, we have
    \begin{align}
        \sum_{i=1}^{t}z_i = & \log\left(\frac{L_t(\MLE{t})}{L_t(\MLE{t-1})} \cdot \frac{L_{t-1}(\MLE{t-1})}{L_{t-1}(\MLE{t-2})}\cdot \cdots \cdot \frac{L_{1}(\MLE{1})}{L_{1}(\MLE{0})}\right) \label{lemma: central limit theorem eq-6} \\
        &= \log\left(\frac{\phi_{t-1}(s_t)^\top\MLE{t} \cdot \phi_{t-2}(s_{t-1})^\top\MLE{t}\cdot\cdots}{\phi_{t-1}(s_t)^\top\MLE{t-1} \cdot \phi_{t-2}(s_{t-1})^\top\MLE{t-2}\cdot\cdots}\right) + \frac{\lambda}{2} \left(\lVert \MLE{t} \rVert_2^2 - \lVert \MLE{0} \rVert_2^2\right)\label{lemma: central limit theorem eq-7}\\
        &= \sum_{i=1}^{t-1} \log{\frac{\phi_{i}(s_{i+1})^\top\MLE{t}}{\phi_{i}(s_{i+1})^\top\MLE{i}}} + \frac{\lambda}{2} \left(\lVert \MLE{t} \rVert_2^2 - \lVert \MLE{0} \rVert_2^2\right)\label{lemma: central limit theorem eq-8} \\
        & \leq \frac{8d^2}{p^2_{\text{min}}}\log\left( \frac{d\lambda+(t-1)L^2}{d}\right) \label{lemma: central limit theorem eq-9},
    \end{align}
    where (\ref{lemma: central limit theorem eq-9}) holds by Lemma \ref{lemma: regret of online learning}.  
    \item For $M_t$, we have
    \begin{align}
        M_t =& \sum_{i=1}^{t} \left(\log{L_{i+1}(\MLE{i})} - \log{L_{i}(\MLE{i-1})} - z_i\right)^2 \label{lemma: central limit theorem eq-10}\\ 
        = & \sum_{i=1}^{t}\left(\log{\frac{\phi_{i}(s_{i+1})^\top\MLE{i}}{\phi_{i}(s_{i+1})^\top\btheta^*}} \right)^2\label{lemma: central limit theorem eq-11} \\
        \leq & \frac{1}{p^2_{\text{min}}} \sum_{i=1}^{t} \lVert \phi_i(s_{i+1}) \rVert^2_{\bA^{-1}_i} \cdot \lVert \MLE{i} - \btheta^*\rVert^2_{\bA_i} \label{lemma: central limit theorem eq-12}  \\
        \leq &  \frac{\max_{i\leq t}\lVert \MLE{i} - \btheta^*\rVert^2_{\bA_i}}{p^2_{\text{min}}} \sum_{i=1}^{t} \lVert \phi_i(s_{i+1}) \rVert^2_{\bA^{-1}_i} \label{lemma: central limit theorem eq-13}\\
        \leq & \max_{i\leq t}\lVert \MLE{i} - \btheta^*\rVert^2_{\bA_i} \cdot \frac{2d}{p^2_{\text{min}}} \cdot \log\left( \frac{d\lambda+tL^2}{d}\right)\label{lemma: central limit theorem eq-14}
    \end{align}
    where (\ref{lemma: central limit theorem eq-12}) holds by Assumption \ref{assumption: p_min} and Cauchy–Schwarz inequality, and (\ref{lemma: central limit theorem eq-14}) holds by Lemma \ref{lemma: summation of vector norm}. 
\end{itemize}
Then, combining (\ref{lemma: central limit theorem eq-5}), (\ref{lemma: central limit theorem eq-9}) and (\ref{lemma: central limit theorem eq-14}) into (\ref{lemma: central limit theorem eq-2}), for all $t$, we have
    \begin{align}
        \lVert \MLE{t} - \btheta^*\rVert^2_{\bA_t} \leq & \frac{1}{p_{\text{min}}} + \frac{8d^2}{p^2_{\text{min}}}\log\left( \frac{d\lambda+tL^2}{d}\right) \nonumber \\
        & + \max_{i\leq t}\lVert \MLE{i} - \btheta^*\rVert_{\bA_{i}} \cdot \sqrt{\frac{4d}{p^2_{\text{min}}} \cdot \log\left( \frac{d\lambda+tL^2}{d}\right)\cdot \log{\frac{1}{\delta}}} \label{lemma: central limit theorem eq-15}
    \end{align}
    holds with probability at least $1-\delta$. Letting $k_t := \argmax_{i\leq t} \lVert \MLE{i} - \btheta^*\rVert_{\bA_i} \leq t$ and denoting the following \hungyh{indicator functions:
    \begin{align}
        & J := \mathbbm{1}\left\{ \lVert \MLE{t} - \btheta^*\rVert^2_{\bA_{t}} \geq \frac{37d^2}{p^2_{\text{min}}}\cdot \log\left( \frac{d\lambda+tL^2}{d}\right)\cdot \log{\frac{1}{\delta}} \right\} \\
        & D_i := \mathbbm{1}\{k_t = i\}, \forall i \leq t,
    \end{align}
    Then, by (\ref{lemma: central limit theorem eq-15}) and the definition of $k_t$}, we have 
    \begin{align}
         \lVert \MLE{t} - \btheta^*\rVert^2_{\bA_{t}} \leq &\; \lVert \MLE{k_t} - \btheta^*\rVert^2_{\bA_{k_t}} \\
         \leq & \;\frac{1}{p_{\text{min}}} + \frac{8d^2}{p^2_{\text{min}}}\log\left( \frac{d\lambda+k_tL^2}{d}\right) \nonumber \\
        & \;+ \lVert \MLE{k_t} - \btheta^*\rVert_{\bA_{k_t}} \cdot \sqrt{\frac{4d}{p^2_{\text{min}}} \cdot \log\left( \frac{d\lambda+k_tL^2}{d}\right)\cdot \log{\frac{1}{\delta}}} \label{lemma: central limit theorem eq-16}
    \end{align}
    holds with probability at least $1-\delta$, which implies 
    \begin{align}
        \lVert \MLE{t} - \btheta^*\rVert^2_{\bA_{t}} \leq \frac{37d^2}{p^2_{\text{min}}}\cdot \log\left( \frac{d\lambda+tL^2}{d}\right)\cdot \log{\frac{1}{\delta}} := \beta_t \label{lemma: central limit theorem eq-17}.
    \end{align}
    By the fact that $\sum_{i=1}^t\text{Pr}(D_i=1) = 1$ and $\text{Pr}(J|D_i=1) \leq \delta, \forall i \in [t]$ shown above, we have
    \begin{align}
        \text{Pr}(J) = \sum_{i=1}^t\text{Pr} (D_i=1)\text{Pr}(J|D_i=1) \leq \delta.
    \end{align}
    We complete the proof.
\end{proof}
\begin{lemma*}
At each time $t$, with probability at least $1-\delta$, we have
\begin{align}
    \nabla_{\btheta}\ell_t(\btheta)|_{\btheta= \MLE{t}}^\top(\MLE{t} - \btheta^*) \leq \beta'_t, \label{lemma: central limit theorem eq-19}
\end{align}
where $\beta'_t:= \frac{22d^2}{p^2_{\text{min}}}\log\left( \frac{d\lambda+tL^2}{d}\right) \cdot \max\{1, \log{\frac{1}{\delta}}\}$.
\end{lemma*}
\begin{proof}
    This lemma can be proved by the same argument as Theorem \ref{lemma: central limit theorem}. By the fact that $\ell_{t+1}(\MLE{t}) - \ell_{t+1}(\btheta^{*}) \geq \nabla_{\btheta}\ell_t(\btheta)|_{\btheta= \MLE{t}}^\top(\btheta^* - \MLE{t}) - \frac{1}{p_{\text{min}}}$, (\ref{lemma: central limit theorem eq-2}), (\ref{lemma: central limit theorem eq-9}), and (\ref{lemma: central limit theorem eq-14}), we also have 
    \begin{align}
        \nabla_{\btheta}\ell_t(\btheta)|_{\btheta= \MLE{t}}^\top(\MLE{t} - \btheta^*) 
        \leq & \frac{1}{p_{\text{min}}} + \frac{8d^2}{p^2_{\text{min}}}\log\left( \frac{d\lambda+k_tL^2}{d\lambda}\right) \nonumber \\
        & + \lVert \MLE{k_t} - \btheta^*\rVert_{\bA_{k_t}} \cdot \sqrt{\frac{4d}{p^2_{\text{min}}} \cdot \log\left( \frac{d\lambda+k_tL^2}{d}\right)\cdot \log{\frac{1}{\delta}}} \label{lemma: central limit theorem eq-18} \\
        \leq & \frac{22d^2}{p^2_{\text{min}}}\log\left( \frac{d\lambda+tL^2}{d}\right) \cdot \max\{1, \log{\frac{1}{\delta}}\} := \beta'_t \label{lemma: central limit theorem eq-19.5}, 
    \end{align}
    where (\ref{lemma: central limit theorem eq-19.5}) holds by plugging (\ref{lemma: central limit theorem eq-17}) into (\ref{lemma: central limit theorem eq-18}).
\end{proof}

\subsection{Proof of Theorem \ref{theorem: upper bound of regret}}
\label{app:thm:regret}
Recalling that 
\begin{align}
    &\beta_t  := \frac{37d^2}{p^2_{\text{min}}}\cdot \log\left( \frac{d\lambda+tL^2}{d}\right)\cdot \log{\frac{1}{\delta}} \\ 
    &\beta'_t = \frac{22d^2}{p^2_{\text{min}}}\log\left( \frac{d\lambda+tL^2}{d}\right) \cdot \max\{1, \log{\frac{1}{\delta}}\},
\end{align}
we then state the detailed form of the regret upper bound.
\begin{theorem*}
    For all linear kernel MDP $\mathcal{M} = \langle \mathcal{S}, \mathcal{A}, P, R, T, \mu_0\rangle$, with probability at least $1-\frac{1}{T} - \delta$, VBMLE, proposed in Algorithm \ref{alg:VBMLE}, has regret upper bound satisfies that 
    \begin{align}
    \normalfont
        \mathcal{R}(T) = & \sum_{t=1}^{T} \left(V^*(s_t,\btheta^*) - V^{\pi_t}(s_t;\btheta^*)\right) \nonumber \\
        \leq & \left(\frac{\beta_T}{2p^2_{\text{min}}} + \beta'_T\right)\cdot \sum_{t=1}^{T}\frac{1}{\alpha(t)} + \frac{4\gamma}{1-\gamma}\sqrt{T\log{\frac{1}{\delta}}} \nonumber \\
        & + \frac{2\gamma}{p_{\text{min}}(1-\gamma)^2}\sqrt{T\log{\frac{1}{\delta}}} + \frac{\sqrt{\beta_T}\gamma}{p_{\text{min}}(1-\gamma)^2} \left(1+ \sqrt{T\cdot\SVN{T}}\right) \nonumber \\
        & + \frac{2d\cdot\alpha(T)}{(1-\gamma)} \cdot \log\left( \frac{d\lambda+TL^2}{d}\right) \label{theorem: upper bound of regret eq-1} . 
    \end{align}
    By choosing $\alpha(t) = \sqrt{t}$, we have $\mathcal{R}(T) = \mathcal{O}(d\sqrt{T}\log{T}/(1-\gamma)^2)$.
\end{theorem*}
\begin{proof}
    By the definition of the cumulative regret in (\ref{eq: def of regret}), we have
    \begin{align}
        \mathcal{R}(T) = & \sum_{t=1}^{T} \left(V^*(s_t,\btheta^*) - V^{\pi_t}(s_t;\btheta^*)\right) \label{theorem: upper bound of regret eq-2}\\
         \leq & \sum_{t=1}^{T} \left( V^*(s_t,\bthetaR) - V^{\pi_t}(s_t;\btheta^*) + \frac{\ell_t(\bthetaR) - \ell_t(\btheta^*)}{\alpha(t)}\right)\label{theorem: upper bound of regret eq-3} \\
         = & \mathcal{R}'(T) + \sum_{t=1}^{T}\frac{\ell_t(\bthetaR) - \ell_t(\btheta^*)}{\alpha(t)}\label{theorem: upper bound of regret eq-3.5}
    \end{align}
    where (\ref{theorem: upper bound of regret eq-3}) holds due to the following inequality:
    \begin{align}
        &\ell_t(\bthetaR) + \alpha(t)V^*(s_t;\bthetaR) \geq \ell_t(\btheta^*) + \alpha(t)V^*(s_t;\btheta^*) \label{theorem: upper bound of regret eq-4} \\
        \implies & V^*(s_t;\btheta^*) \leq V^*(s_t;\bthetaR) + \frac{\ell_t(\bthetaR) - \ell_t(\btheta^*)} {\alpha(t)} \label{theorem: upper bound of regret eq-5},
    \end{align}
    and (\ref{theorem: upper bound of regret eq-3.5}) holds by $\mathcal{R}'(T) := \sum^T_{t=1}(V^*(s_t,\bthetaR) - V^{\pi_t}(s_t;\btheta^*))$. For the term $\sum_{t=1}^{T}(\ell_t(\bthetaR) - \ell_t(\btheta^*))/\alpha(t)$, we have
    \begin{align}
        \sum_{t=1}^{T} \frac{\ell_t(\bthetaR) - \ell_t(\btheta^*)}{\alpha(t)} = & \sum_{t=1}^{T}\frac{\ell_t(\bthetaR) - \ell_t(\MLE{t})}{\alpha(t)} + \sum_{t=1}^{T}\frac{\ell_t(\MLE{t}) - \ell_t(\btheta^*)}{\alpha(t)} \label{theorem: upper bound of regret eq-6} \\
        \leq &  -\sum_{t=1}^{T} \frac{1}{2\alpha(t)} \lVert \bthetaR - \MLE{t} \rVert^2_{\bA_t(\btheta')} + \sum_{t=1}^{T}\frac{1}{2\alpha(t)} \lVert \MLE{t} - \btheta^* \rVert^2_{\bA_t(\btheta'')}\nonumber \\
        & - \sum_{t=1}^{T}\frac{\nabla_{\btheta}\ell_t(\btheta)|_{\btheta = \MLE{t}}^\top(\btheta^*-\MLE{t})}{\alpha(t)} \label{theorem: upper bound of regret eq-7}\\ 
        \leq & \sum_{t=1}^{T}\frac{1}{2p^2_{\text{min}}\alpha(t)} \lVert \MLE{t} - \btheta^* \rVert^2_{\bA_t} - \sum_{t=1}^{T}\frac{\nabla_{\btheta}\ell_t(\btheta)|_{\btheta = \MLE{t}}^\top(\btheta^*-\MLE{t})}{\alpha(t)} \nonumber \\
        & - \sum_{t=1}^{T}\frac{1}{2\alpha(t)} \lVert \bthetaR - \MLE{t} \rVert^2_{\bA_t}\label{theorem: upper bound of regret eq-8} \\
        \leq & \left(\frac{\beta_T}{2p^2_{\text{min}}} + \beta'_T\right)\cdot \sum_{t=1}^{T}\frac{1}{\alpha(t)} - \sum_{t=1}^{T}\frac{1}{2\alpha(t)} \lVert \bthetaR - \MLE{t} \rVert^2_{\bA_t(\btheta')}, \label{theorem: upper bound of regret eq-9}
    \end{align}
    where (\ref{theorem: upper bound of regret eq-7}) holds by applying Taylor's theorem with $\btheta' \in (\bthetaR, \MLE{t}), \btheta''\in(\MLE{t},\btheta^*)$,  and the fact that $\nabla_{\btheta}\ell_t(\btheta)|_{\btheta = \MLE{t}}^\top(\bthetaR-\MLE{t}) \leq 0$, (\ref{theorem: upper bound of regret eq-8}) holds due to $-p^2_{\text{min}}\nabla^2_{\btheta}\ell_t(\btheta) \preceq  \bA_{t} \preceq -\nabla^2_{\btheta}\ell_t(\btheta), \forall \btheta \in \mathbb{P}$, and (\ref{theorem: upper bound of regret eq-9}) holds with probability at least $1-\sum_{t=1}^{T}\frac{1}{T^2}$ by (\ref{lemma: central limit theorem eq-17}), (\ref{lemma: central limit theorem eq-19}), and replacing $\delta$ with $\frac{1}{T^2}$ in $\beta_T$ and $\beta'_T$, Lemma \ref{lemma: summation of vector norm}, and Cauchy–Schwarz inequality. Then, we have
    \begin{align}
        \mathcal{R}'(T) = &\; \sum_{t=1}^{T}\left[ V^*(s_t,\bthetaR) - V^{\pi_t} (s_t;\btheta^*)\right] \label{theorem: upper bound of regret eq-11}\\
        = &\; \gamma \sum_{t=2}^{T+1}\left[ \mathbb{E}_{s'\sim \text{Pr}(\cdot|s_t,a_t;\bthetaR)}[V^*(s',\bthetaR)]  - \mathbb{E}_{s'\sim \text{Pr}(\cdot|s_t,a_t;\btheta^*)}[V^{\pi_t}(s',\btheta^*)]\right] \label{theorem: upper bound of regret eq-12} \\
        = &\; \gamma \sum_{t=2}^{T+1}\bigg[ \underbrace{\mathbb{E}_{s'\sim \text{Pr}(\cdot|s_t,a_t;\btheta^*)}[{\color{black}V^*(s',\bthetaR)} - V^{\pi_t}(s',\btheta^*)] - {\color{black}\left( V^*(s_{t+1},\bthetaR) - V^{\pi_t}(s_{t+1},\btheta^*)\right)}}_{:=B_1} \nonumber \\
        &\; + \underbrace{\mathbb{E}_{s'\sim \text{Pr}(\cdot|s_t,a_t;\bthetaR)}[V^*(s',\bthetaR)] {\color{black} - V^{\pi_t}(s_{t+1},\btheta^*)}}_{:=B_2}\nonumber \\
        &\; \underbrace{-\mathbb{E}_{s'\sim \text{Pr}(\cdot|s_t,a_t;\btheta^*)}[{\color{black}V^*(s',\bthetaR)}] {\color{black} + V^*(s_{t+1},\bthetaR) }}_{:=B_3} \bigg] \label{theorem: upper bound of regret eq-13}\\
        \leq & \frac{4\gamma}{1-\gamma}\sqrt{T\log{\frac{1}{\delta}}} + B_2 \label{theorem: upper bound of regret eq-14}
    \end{align}
    where (\ref{theorem: upper bound of regret eq-12}) holds by $\pi_t = \argmax_{\pi}V^{\pi}(s_t,\bthetaR)$, and (\ref{theorem: upper bound of regret eq-14}) holds with probability at least $1-\sum_{t=1}^{T} \frac{2}{t^2}$ by applying Azuma Hoeffding inequality in Lemma \ref{lemma: Azuma–Hoeffding inequality} on $B_1$ and $B_3$, which are martingale difference sequences. For $B_2$, we have
    \begin{align}
        &\gamma \sum_{t=2}^{T+1} \left[\mathbb{E}_{s'\sim \text{Pr}(\cdot|s_t,a_t;\bthetaR)}[V^*(s',\bthetaR)]  - V^{\pi_t}(s_{t+1},\btheta^*)\right] \nonumber\\
        = & \gamma \sum_{t=2}^{T+1} \left[\mathbb{E}_{s'\sim \text{Pr}(\cdot|s_t,a_t;{\color{black}\btheta^*})}\left[{\color{black}\frac{\text{Pr}(s'|s_t,a_t;\bthetaR)}{\text{Pr}(s'|s_t,a_t;\btheta^*)}}V^*(s',\bthetaR)\right] - V^{\pi_t}(s_{t+1},\btheta^*)\right]\label{theorem: upper bound of regret eq-15} \\
        = & \underbrace{\gamma \sum_{t=2}^{T+1} \left[\mathbb{E}_{s'\sim \text{Pr}(\cdot|s_t,a_t;\btheta^*)}\left[\frac{\text{Pr}(s'|s_t,a_t;\bthetaR)}{\text{Pr}(s'|s_t,a_t;\btheta^*)}V^*(s',\bthetaR)\right] {\color{black} - \frac{\text{Pr}(s_{t+1}|s_t,a_t;\bthetaR)}{\text{Pr}(s_{t+1}|s_t,a_t;\btheta^*)} V^*(s_{t+1},\bthetaR)}\right]}_{:=B_4} \nonumber \\
        & + \gamma \sum_{t=2}^{T+1} \left[{\color{black}\frac{\text{Pr}(s_{t+1}|s_t,a_t;\bthetaR)}{\text{Pr}(s_{t+1}|s_t,a_t;\btheta^*)}V^*(s_{t+1},\bthetaR)}-V^{\pi_t}(s_{t+1},\btheta^*)\right] \label{theorem: upper bound of regret eq-16}\\
        = & B_4 + \gamma\mathcal{R}'(T) + \frac{2\gamma}{1-\gamma}+ \gamma \sum_{t=2}^{T+1} \left[\left(\frac{\text{Pr}(s_{t+1}|s_t,a_t;\bthetaR)}{\text{Pr}(s_{t+1}|s_t,a_t;\btheta^*)}-1\right)V^*(s_{t+1},\bthetaR)\right]\label{theorem: upper bound of regret eq-17} \\
        \leq &\frac{2\gamma}{p_{\text{min}}(1-\gamma)}\sqrt{T\log{\frac{1}{\delta}}} + \underbrace{\gamma \sum_{t=2}^{T+1} \left[\left(\frac{\text{Pr}(s_{t+1}|s_t,a_t;\bthetaR)}{\text{Pr}(s_{t+1}|s_t,a_t;\btheta^*)}-1\right)V^*(s_{t+1},\bthetaR)\right]}_{:=B_5} + \gamma\mathcal{R}'(T) + \frac{2\gamma}{1-\gamma}\label{theorem: upper bound of regret eq-18} 
    \end{align}
    where (\ref{theorem: upper bound of regret eq-15}) holds by importance sampling, (\ref{theorem: upper bound of regret eq-16}) holds by adding and subtracting ${\color{black}\text{Pr}(s_{t+1}|s_t,a_t;\bthetaR)/\text{Pr}(s_{t+1}|s_t,a_t;\btheta^*)}\cdot V^*(s_{t+1},\bthetaR)$, (\ref{theorem: upper bound of regret eq-17}) 
    holds by Lemma \ref{Lemma: upper bound of V}, and (\ref{theorem: upper bound of regret eq-18}) holds with probability at least $1-\delta$ by applying Azuma-Hoeffding inequality in Lemma \ref{lemma: Azuma–Hoeffding inequality} on $B_4$. 
    Then, for the term $B_5$, we have
    \begin{align}
        B_5 = &\gamma \sum_{t=2}^{T+1} \left[\left(\frac{\text{Pr}(s_{t+1}|s_t,a_t;\bthetaR)}{\text{Pr}(s_{t+1}|s_t,a_t;\btheta^*)}-1\right)V^*(s_{t+1},\bthetaR)\right] \label{theorem: upper bound of regret eq-19} \\
        \leq & \frac{\gamma}{p_{\text{min}}(1-\gamma)} \sum_{t=1}^{T} |\innerprod{\phi(s_{t+1}|s_t,a_t)}{\bthetaR-\btheta^*}| + \frac{\gamma}{p_{\text{min}}(1-\gamma)}\label{theorem: upper bound of regret eq-20} \\
        \leq & \frac{\gamma}{p_{\text{min}}(1-\gamma)} \sum_{t=1}^{T} \lVert\phi(s_{t+1}|s_t,a_t)\rVert_{\bA_{t}^{-1}}\cdot\lVert\bthetaR-\MLE{t}\rVert_{\bA_t}  \nonumber \\
        & + \frac{\gamma}{p_{\text{min}}(1-\gamma)} \sum_{t=1}^{T} \lVert\phi(s_{t+1}|s_t,a_t)\rVert_{\bA_{t}^{-1}}\cdot\lVert\MLE{t}-\btheta^*\rVert_{\bA_t} + \frac{\gamma}{p_{\text{min}}(1-\gamma)} \label{theorem: upper bound of regret eq-21} \\
        \leq & \frac{\gamma}{p_{\text{min}}(1-\gamma)} \sum_{t=1}^{T} \lVert\phi(s_{t+1}|s_t,a_t)\rVert_{\bA_{t}^{-1}}\cdot\lVert\bthetaR-\MLE{t}\rVert_{\bA_t} \nonumber \\ 
        & + \frac{\sqrt{\beta_T}\gamma}{p_{\text{min}}(1-\gamma)} \left(1+ \sqrt{T\cdot\SVN{T}}\right) \label{theorem: upper bound of regret eq-22}
    \end{align}
    where (\ref{theorem: upper bound of regret eq-20}) holds by $p_{\text{min}} \leq \min_{t\leq T}\text{Pr}(s_{t+1}|s_t,a_t;\btheta^*)$, which is defined in Assumption \ref{assumption: p_min},  (\ref{theorem: upper bound of regret eq-21}) holds by Cauchy–Schwarz inequality and triangle inequality, and (\ref{theorem: upper bound of regret eq-22}) holds by Theorem \ref{lemma: central limit theorem} with probability at least $1-\frac{1}{T}$ by replacing $\delta$ with $\frac{1}{T^2}$ in $\beta_T$ and Lemma \ref{lemma: summation of vector norm}. Combining the final term in (\ref{theorem: upper bound of regret eq-9}) and (\ref{theorem: upper bound of regret eq-22}), we have
    \begin{align}
        &  \sum_{t=1}^{T} \left(- \frac{1}{2\alpha(t)} \lVert \bthetaR - \MLE{t} \rVert^2_{\bA_t} + \frac{\gamma}{p_{\text{min}}(1-\gamma)}\lVert\phi(s_{t+1}|s_t,a_t)\rVert_{\bA_{t}^{-1}}\lVert\bthetaR-\MLE{t}\rVert_{\bA_t} \right)\nonumber \\
        \leq &  \left(\frac{\gamma}{4p_{\text{min}}(1-\gamma)}\right)^2\sum_{t=1}^{T} \alpha(t)\lVert\phi(s_{t+1}|s_t,a_t)\rVert^2_{\bA_{t}^{-1}} \label{theorem: upper bound of regret eq-23} \\
        = & \alpha(T) \cdot \SVN{T}, \label{theorem: upper bound of regret eq-24} 
    \end{align}
    where (\ref{theorem: upper bound of regret eq-23}) holds by completing the square, and (\ref{theorem: upper bound of regret eq-24}) holds by Lemma \ref{lemma: summation of vector norm}. Letting $\alpha(t) = \sqrt{t}$ and combining (\ref{theorem: upper bound of regret eq-3.5}), (\ref{theorem: upper bound of regret eq-9}), (\ref{theorem: upper bound of regret eq-14}), (\ref{theorem: upper bound of regret eq-18}), 
    (\ref{theorem: upper bound of regret eq-22}), and (\ref{theorem: upper bound of regret eq-24}), we complete the proof.
\end{proof}

\section{Implementation Details}
\subsection{Environment}
\label{section: env}
\begin{itemize}
    \item $\phi(\cdot|s, a)$: We employ a neural network architecture with two linear hidden layers, each utilizing the Rectified Linear Unit (ReLU) activation function applied to every neuron. The network's input is created by concatenating the one-hot vectors derived from the state and action indices. The output of this network is subsequently transformed into a set of $d$ final layers, each having a dimension of $hidden\_size \times |\mathcal{S}|$, and employing the softmax activation function. The resulting outputs from these final layers are concatenated to yield the final output, which can be represented as $\{\phi(s'|s, a)\}_{s' \in \mathcal{S}} \in \mathbb{R}^{d\times|\mathcal{S}|}$.
    \item $\btheta^*$: By initializing the parameter vector $\btheta^*$ with random values such that the summation of its elements equals 1, we can readily verify that the $\btheta^*$ resides within the probability simplex.
\end{itemize}
All the simulations are conducted on the device with (i) CPU: Intel Core i7-11700K, (ii) RAM: 32 GB, (iii) GPU: RTX 3080Ti, and (iv) OS: Windows 10.
\subsection{Hyper-parameters}
\label{section: hyperparameter}
\begin{table}[!ht]
\centering
\begin{tabular}{|c|c|}
\hline
$\gamma$                            & $0.9$  \\ \hline
Temperature of the softmax function & $0.01$ \\ \hline
$\lambda$                           & $1$    \\ \hline
$\delta$ for UCLK                    & $0.1$  \\ \hline
\end{tabular}
\end{table}

\section{Additional Simulation Result}
Figure \ref{fig: S=15, A=4} shows the regret performance of VBMLE under a larger linear MDP, with $|\mathcal{S}| = 15, |\mathcal{A}| = 4$. We don't provide the result of UCLK since it is impractical in terms of computation time. The result includes two variants of the biased term designed in VBMLE:
\begin{itemize}
    \item \textbf{Approximated VBMLE:} The biased term equals to $\sum_{s'\in\mathcal{S}} \innerprod{\phi(s'|s_t,a_t)V^*(s';\btheta^{\textbf{V}}_{t-1})}{\btheta}$. Notice that the term $V^*(s';\btheta^{\textbf{V}}_{t-1})$ is detached from $\btheta$.
    \item \textbf{Exact VBMLE:} The biased term $V^*(s_t;\btheta)$ is constructed by value iteration (\ref{alg:vi}), which is the same implementation as that in Figure \ref{fig:regret}.
\end{itemize}
\subsection{Standard Deviation of Figure \ref{fig:regret}}
\begin{table}[!ht]
\label{table: std of regret}
\centering
\caption{The table shows the standard deviation of cumulative regret at $t=500$.}
\begin{tabular}{|c|c|c|}
\hline
               & $|\mathcal{S}| = 3, |\mathcal{A}| = 2$ & $|\mathcal{S}| = 5, |\mathcal{A}| = 4$  \\ \hline
\textbf{UCLK}  & $79.289$                               & $23.982$                                                                    \\ \hline
\textbf{VBMLE} & $1.874$                                & $1.830$                                                                 \\ \hline
\end{tabular}
\end{table}

\begin{algorithm}
    \caption{Value Iteration}
    \begin{algorithmic}[1]
        \State {\bfseries Input:} $\delta,U,\btheta$
        \State $V^{(1)}(\cdot;\btheta) = \frac{1}{1-\gamma}$
        \For{$u=1,2,\cdots,U$}
            \State $V^{(u+1)}(\cdot;\btheta) = \max_a \left\{R(\cdot,a) + \gamma \sum\limits_{s'\in \mathcal{S}}P(s'|s,a;\btheta)V^{(u)}(\cdot;\btheta)\right\}$
        \EndFor
        \State Return $V^{(U+1)}(\cdot;\btheta)$
    \end{algorithmic}
    \label{alg:vi}
\end{algorithm}

\begin{figure}[!ht]
\centering
$\begin{array}{c}
    \multicolumn{1}{l}{\mbox{\bf }} \\
    \scalebox{0.49}{\includegraphics[width=\textwidth]{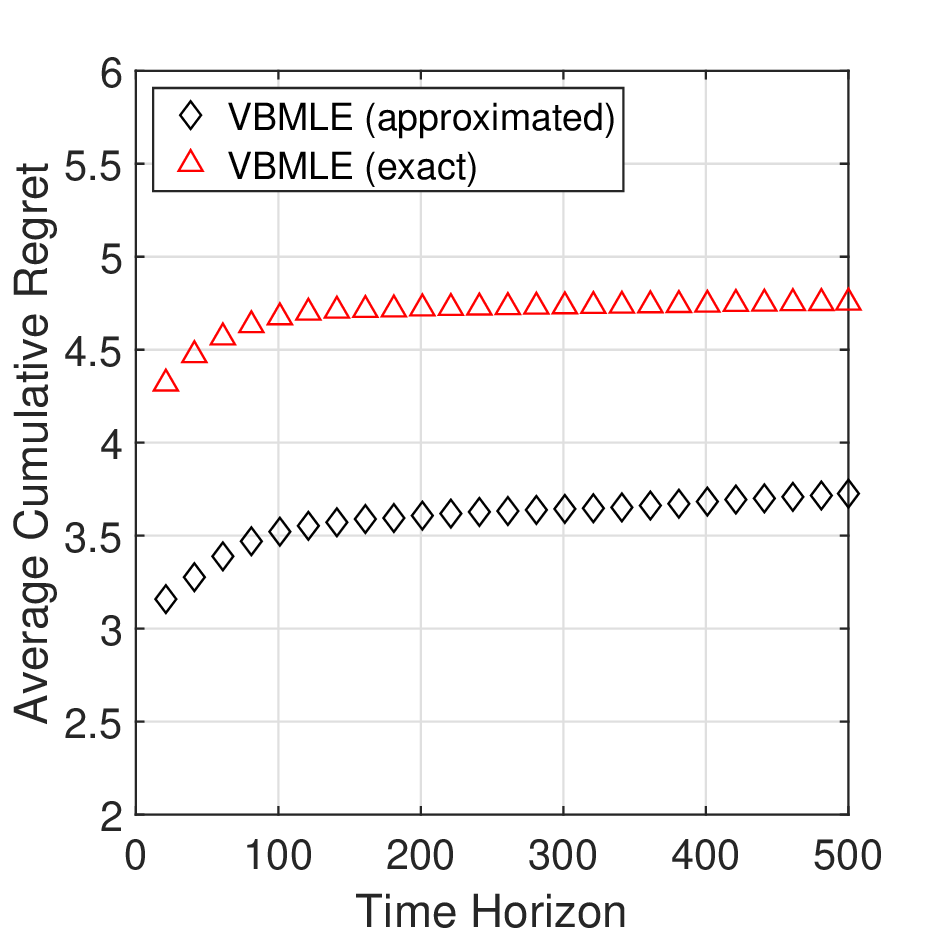}}  \\
    \mbox{(a) $|\mathcal{S}| = 15, |\mathcal{A}| = 4$} 
\end{array}$
\label{fig: S=15, A=4}
\caption{Regret averaged over $20$ trials with $T = 500$}
\end{figure}

\end{document}